\documentclass[journal]{IEEEtran}
\usepackage{amsmath,amsfonts}
\usepackage{algorithmic}
\usepackage[colorlinks, linkcolor=red, anchorcolor=blue, citecolor=green]{hyperref}
\usepackage[capitalize]{cleveref}
\usepackage{array}
\usepackage[caption=false,font=normalsize,labelfont=sf,textfont=sf]{subfig}
\usepackage{textcomp}
\usepackage{stfloats}
\usepackage{verbatim}
\usepackage{graphicx}
\usepackage{cite}
\usepackage{color,xcolor}
\usepackage{booktabs}
\usepackage{multirow}
\usepackage{pifont}
\newcommand{\etal}{\textit{et al.}}
\newcommand{\nbf}[1]{\noindent \textbf{#1.}}

\newcommand{\pexp}{\noindent where }

\crefname{section}{Sec.}{Sec.}
\crefname{table}{Table}{Tables}
\crefname{figure}{Figure}{Figures}

\def\BibTeX{{\rm B\kern-.05em{\sc i\kern-.025em b}\kern-.08em
    T\kern-.1667em\lower.7ex\hbox{E}\kern-.125emX}}
\usepackage{balance}

\begin{document}

\title{Spectrum-driven Mixed-frequency Network \\ for Hyperspectral Salient Object Detection}

\author{Peifu Liu, Tingfa Xu$^{\dagger}$, Huan Chen, Shiyun Zhou, Haolin Qin, Jianan Li$^{\dagger}$
\thanks{Peifu Liu, Tingfa Xu, Huan Chen, Shiyun Zhou, Haolin Qin, and Jianan Li are with Beijing Institute of Technology, Beijing 10081, China. Email: \{laprf, ciom\_xtf1, huanchen, zhoushiyun, 3120225333, lijianan\}@bit.edu.cn} %
\thanks{Tingfa Xu is also with the Key Laboratory of Photoelectronic Imaging Technology and System, Ministry of Education of China, Beijing 100081, China, and with the Chongqing Innovation Center, Chongqing 401135, China.
} %
\thanks{$^{\dagger}$ Correspondence to: Tingfa Xu and Jianan Li.}
}%

\markboth{Journal of \LaTeX\ Class Files,~Vol.~18, No.~9, September~2020}%
{How to Use the IEEEtran \LaTeX \ Templates}

\maketitle

\begin{abstract}
Hyperspectral salient object detection (HSOD) aims to detect spectrally salient objects in hyperspectral images (HSIs). However, existing methods inadequately utilize spectral information by either converting HSIs into false-color images or converging neural networks with clustering. We propose a novel approach that fully leverages the spectral characteristics by extracting two distinct frequency components from the spectrum: low-frequency Spectral Saliency and high-frequency Spectral Edge. The Spectral Saliency approximates the region of salient objects, while the Spectral Edge captures edge information of salient objects. These two complementary components, crucial for HSOD, are derived by computing from the inter-layer spectral angular distance of the Gaussian pyramid and the intra-neighborhood spectral angular gradients, respectively. To effectively utilize this dual-frequency information, we introduce a novel lightweight Spectrum-driven Mixed-frequency Network (SMN). SMN incorporates two parameter-free plug-and-play operators, namely Spectral Saliency Generator and Spectral Edge Operator, to extract the Spectral Saliency and Spectral Edge components from the input HSI independently. Subsequently, the Mixed-frequency Attention module, comprised of two frequency-dependent heads, intelligently combines the embedded features of edge and saliency information, resulting in a mixed-frequency feature representation. Furthermore, a saliency-edge-aware decoder progressively scales up the mixed-frequency feature while preserving rich detail and saliency information for accurate salient object prediction. Extensive experiments conducted on the HS-SOD benchmark and our custom dataset HSOD-BIT demonstrate that our SMN outperforms state-of-the-art methods regarding HSOD performance. Code and dataset will be available at \href{https://github.com/laprf/SMN}{https://github.com/laprf/SMN}.
\end{abstract}

\begin{IEEEkeywords}
Hyperspectral salient object detection, Spectrum, Mixed-frequency Attention
\end{IEEEkeywords}

\section{Introduction}
\IEEEPARstart{H}{yperspectral} imaging systems offer a unique capability to capture data from observed scenes, providing both high spatial resolution and abundant spectral information. This enables the acquisition of hyperspectral images (HSIs) consisting of numerous contiguous narrow spectral bands~\cite{huang2021salient}. By selecting an arbitrary point from the spatial dimension of the hyperspectral cube, the spectral response curve effectively represents the distinctive characteristics of a target. This characteristic has led to the increasing significance of HSIs in various disciplines, including target detection~\cite{cao2015salient}, spectral estimation~\cite{arad2016sparse}, and remote sensing~\cite{remote_sensing_1}.

\begin{figure}
    \centering
    \includegraphics[width=\linewidth]{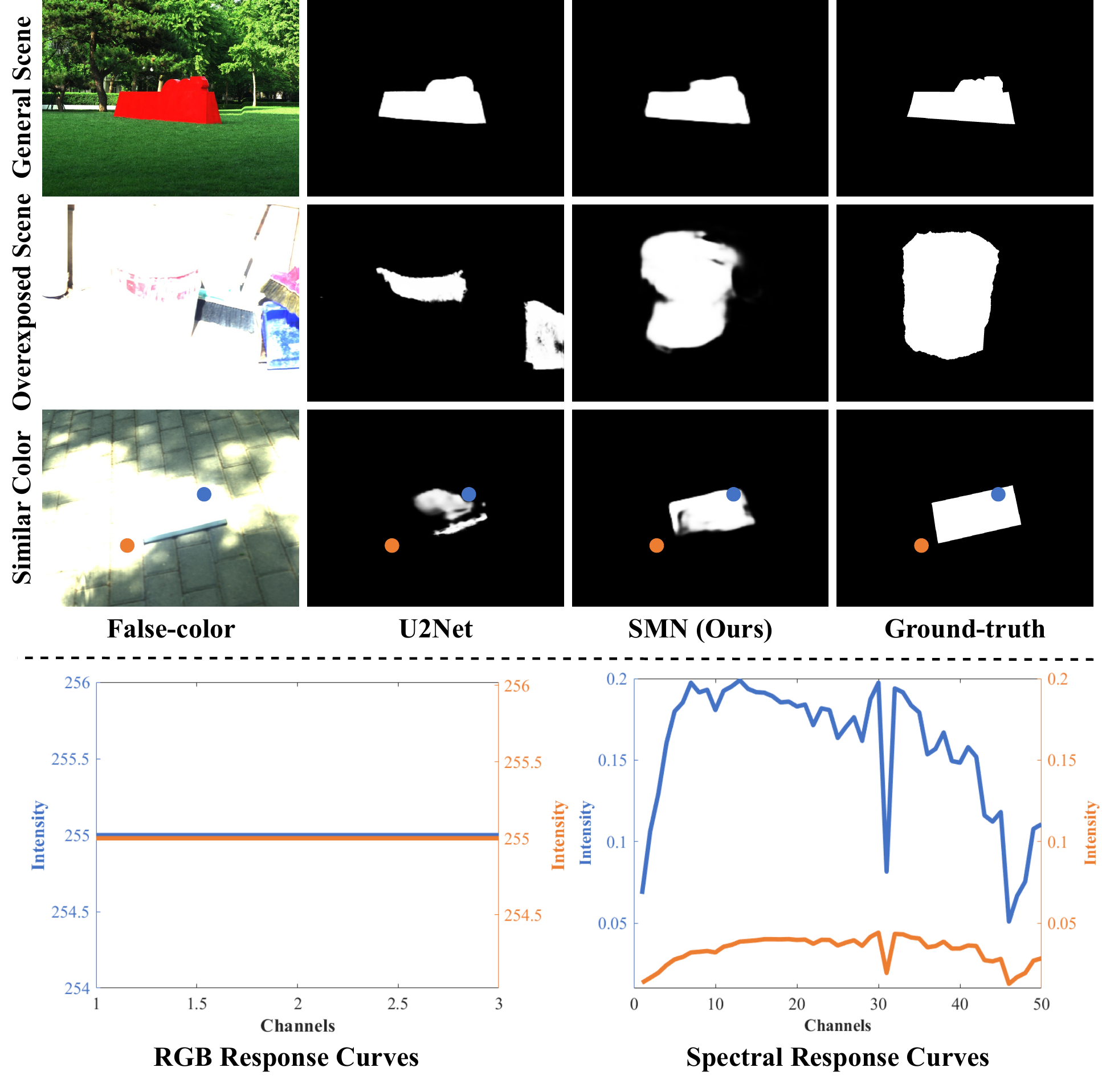}
        \caption{Comparison with an RGB-based saliency detection method, U2Net~\cite{Qin_2020_U2Net}. The input of U2Net is false-color images. We compare the RGB and spectral response curves of the foreground point (blue) and background point (orange). The RGB response remains consistent across all three channels, leading to detection failure. However, the spectral curves of these two points exhibit distinct characteristics, enabling their discrimination using spectrum information. Consequently, our proposed Spectrum-driven Mixed-frequency Network (SMN) outperforms U2Net, particularly in the latter two scenarios.}
    \label{fig1}
\end{figure}

By simulating human visual attention, salient object detection (SOD) aims to locate and segment the most salient object or region in a scene~\cite{borji_2015_what}. Recent advancements in SOD have witnessed the utilization of convolutional neural networks~\cite{li2015visual,zhang2017deep} or Transformers~\cite{liu2021visual,zhang2021learning}, which have significantly enhanced the representation capability of features, resulting in notable improvements in detection performance~\cite{zhuge2022salient}.
However, traditional SOD algorithms heavily rely on color information to discriminate foreground and background objects. In certain exceptional cases, such as scenes with overexposure or situations where the foreground and background colors are similar, the color information of the foreground object may be absent or indistinguishable from the background, leading to detection failures (as illustrated in \cref{fig1}).

In contrast to RGB images, HSIs offer a wealth of spectral information, enabling a more comprehensive characterization of an object's material, composition, and other intrinsic properties. HSIs exhibit a higher resilience to variations in illumination conditions and are less reliant on color and texture information typically present in RGB images. For instance, as depicted in \cref{fig1}, we have selected a foreground point (blue) and a background point (orange) and plotted their respective spectral and RGB response curves. The bottom section of \cref{fig1} clearly illustrates that the RGB response remains consistent across all three channels, making it challenging for false-color images to differentiate between the two points effectively. However, the spectral curves of these two points exhibit distinct characteristics, enabling their discrimination using spectrum information.
Hyperspectral salient object detection (HSOD), which involves detecting the most salient object within HSIs, therefore holds tremendous potential for diverse applications, including pest control~\cite{Liang_2018_material}, military surveillance~\cite{Sun_2022_hyperspectral}, and environmental management~\cite{env_control}.

Traditional methods~\cite{itti1998model, liang2013salient, moan2011saliency} used for HSOD often rely on a simple conversion of HSIs into false-color images, which only exploit a fraction of the available spectral data. Recently, some deep learning techniques~\cite{imamouglu2019salient,huang2021salient} directly employ convolutional neural networks (CNNs) to extract features of entire HSIs and subsequently utilize clustering algorithms to generate saliency maps. Although these methods have achieved remarkable results, they usually encounter several limitations: i) Clustering algorithms are often time-consuming; ii) CNNs extract a large number of features, not all of which are useful for saliency detection, leading to computational redundancy and further burdening time resources; iii) The applicability of clustering algorithms on CNN-extracted features is limited, resulting in suboptimal detection performance compared to end-to-end networks.

To address the above issues, we conducted an extensive investigation of HSIs and identified two distinct frequency components that can be extracted based on the spectrum, playing a crucial role in the accurate identification and localization of salient objects. Specifically, the lower-frequency component, known as Spectral Saliency, provides an approximate localization of the salient target, while the higher-frequency component, known as Spectral Edge, enhances the edges of the target.
Motivated by the aim of fully leveraging these two types of spectrum-extracted information that complement each other, we introduce the Spectrum-driven Mixed-frequency Network (SMN), the first end-to-end deep network designed for HSOD. SMN is a lightweight model, which comprises four key components: the extraction of Spectral Saliency and Spectral Edge maps, frequency-specific embeddings, a Mixed-frequency Attention module, and a decoder aware of both saliency and edge information.

\textbf{Firstly}, we extract the Spectral Saliency and Spectral Edge maps from the input HSIs using two plug-and-play operators: the Spectral Saliency Generator (SSG) and the Spectral Edge Operator (SEO). 
In the SSG, we construct a spatially blurred Gaussian pyramid, with each layer containing the complete spectrum of the input. The spectral angular distance (SAD) between the pyramid layers is utilized to compute Spectral Saliency maps. 
In the SEO, we first calculate the SAD between each pixel and its neighborhood. Subsequently, we employ various kernels to compute the gradient of the SAD values within the neighborhood, enabling the determination of the Spectral Edge maps. 
It is worth noting that these two operators perform their computations rapidly without the need for learnable parameters.

\textbf{Secondly}, Spectral Saliency and Spectral Edge images are subsequently transformed into deep features employing frequency-specified embeddings. 
By leveraging a low-frequency embedding process, deep saliency features are obtained. 
For the high-frequency embedding, we employ cascaded convolutional layers along with an Edge Detection Module, which combines to capture intricate edge details. 
Both low-frequency and high-frequency embeddings generate deep features related to saliency and edge, respectively. 
Considering that edge features are considered low-level features, the high-frequency embedding is designed to be shallower than the low-frequency embedding. This not only enhances the efficiency of the feature extraction process but also minimizes the number of required parameters.

\textbf{Thirdly}, we introduce a Mixed-Frequency Attention module designed to enable intricate interactions and fusion among multi-frequency deep features, resulting in a nuanced mixed-frequency feature representation. 
The module comprises two frequency-dependent heads: a low-frequency head and a high-frequency head.
The low-frequency head ingests saliency features and employs self-attention mechanisms to accentuate regions of importance.
Conversely, the high-frequency head processes both saliency and edge features, executing cross-attention between them. Given that the low-frequency saliency representation is refined in light of neighboring high-frequency edge details, and that unrestricted long-range interactions between frequencies could introduce noise or be counterproductive, we opt for a localized attention paradigm-specifically, the neighborhood attention mechanism-as elaborated in NAT~\cite{Hassani_2023_NAT}. This approach not only ensures contextually relevant interactions within a confined neighborhood but also mitigates computational burden by maintaining linear complexity, unlike the quadratic complexity inherent in traditional attention mechanisms. The synthesized mixed-frequency feature encapsulates a rich amalgamation of both edge and saliency attributes.

\textbf{Finally}, we employ a saliency-edge-aware decoder that progressively upscales the mixed-frequency feature. Since the shallow saliency information and edge details from the frequency-specified embedding phase are simultaneously preserved, a saliency map exhibits high-fidelity edges and superior detection accuracy can be obtained. Consequently, the resulting saliency maps exhibit superior detection performance while minimizing computational costs.

We conducted extensive evaluations of our model on the HS-SOD dataset~\cite{imamoglu2018hyperspectral} as well as our collected dataset, HSOD-BIT. The results demonstrate that our proposed method surpasses the existing state-of-the-art HSOD methods. Our model is lightweight yet capable of detecting salient objects more comprehensively compared to RGB-based SOD methods, particularly in scenarios with overexposure and similar foreground and background colors. 

Our contributions can be summarized as follows:
\begin{itemize}
\item We propose a novel Spectrum-driven Mixed-frequency Network for the task of HSOD. This approach effectively utilizes both low-frequency and high-frequency information present in HSIs to detect salient objects. To our knowledge, our work represents the first attempt to apply an end-to-end neural network to the HSOD problem.
\item We introduce two parameter-free plug-and-play operators, namely the Spectral Saliency Generator and the Spectral Edge Operator, specifically designed for HSIs. These operators enable us to leverage spectral information and provide frequency-specific information effectively.
\item We tailor a Mixed-frequency Attention module to fully exploit the distinct frequency properties present in HSIs. The design of frequency-dependent heads enables the network to focus on different types of information.
\item We present quantitative and qualitative experimental results that demonstrate the superiority of our method compared to state-of-the-art HSOD methods on both the HS-SOD and HSOD-BIT datasets.
\end{itemize}

\section{Related Work}

\subsection{Salient Object Detection}
Conventional SOD methods rely on hand-crafted features~\cite{itti1998model, rosin2009simple, alexe2010object}. For instance, Rosin \etal\cite{rosin2009simple} employ edge detection, threshold decomposition, and pixel-wise operations to identify salient objects. Alexe \etal\cite{alexe2010object} utilize a generic objectness prior by leveraging object proposals. While these hand-crafted features enable real-time SOD, they have several limitations in capturing salient objects in complex scenarios~\cite{zhang2020weakly}. For instance, they tend to emphasize high-contrast edges rather than the salient object itself, and the preservation of boundaries is often inadequate~\cite{sod_survey}.

In contrast, CNNs possess exceptional feature extraction capabilities and can identify the most salient regions without relying on prior knowledge~\cite{li2015visual,zhang2017deep}. Li \etal~\cite{li2015visual} separate the encoding of low-level and high-level information, flatten and concatenate them, and then input the resulting data into a two-layer perceptron to predict the saliency region. Zhang \etal~\cite{zhang2017deep} employ saliency cues and a multi-level fusion mechanism to detect salient objects.
Yao \etal~\cite{Yao_2022_Boundary} integrate the edge extraction module with the prediction network, yielding saliency maps with precise edge delineation.
However, these SOD techniques are limited to RGB data and cannot be directly applied to hyperspectral data for HSOD.

\begin{figure*}
    \centering
    \includegraphics[width=\linewidth]{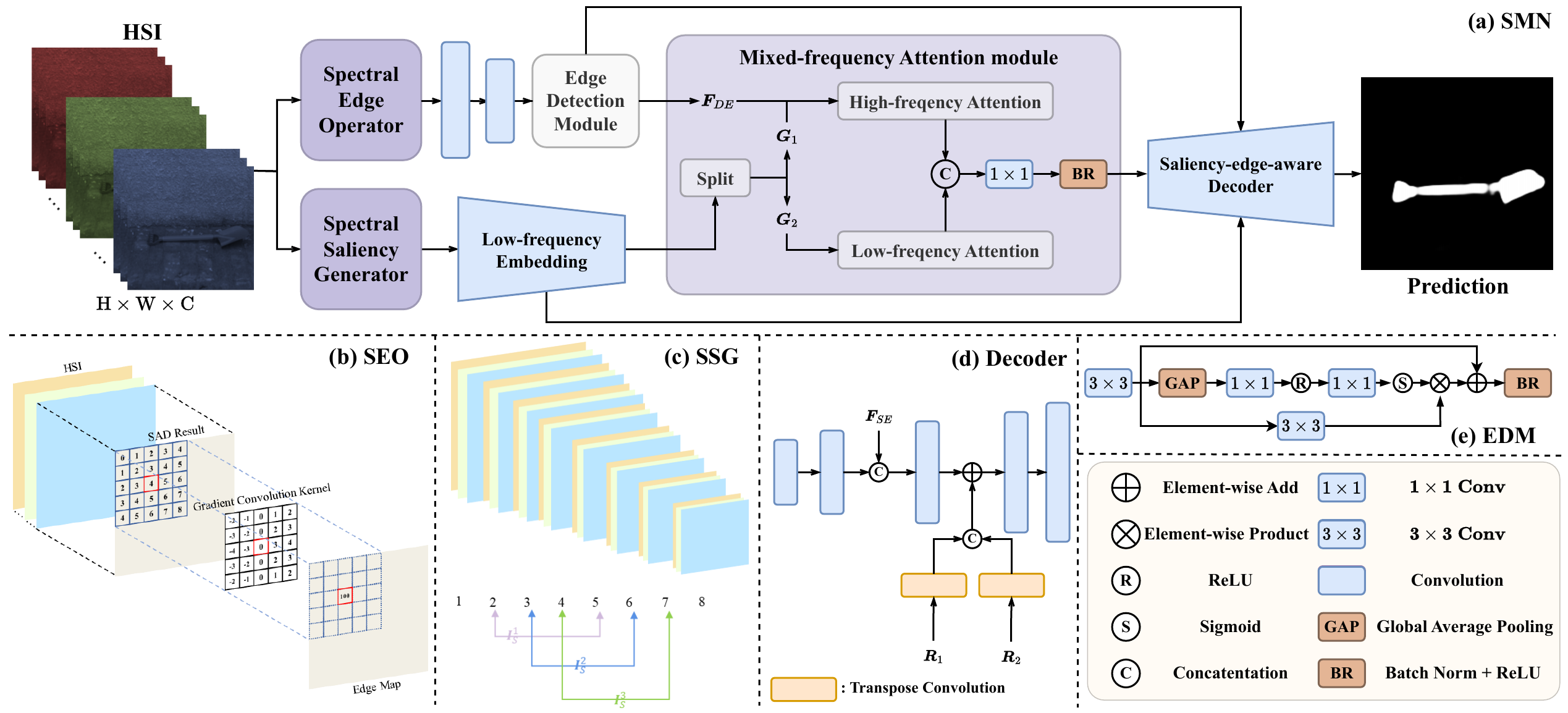}
        \caption {(a) Illustration of the Spectrum-Driven Mixed-Frequency Network (SMN) employing an encoder-bottleneck-decoder architecture. The encoder comprises two distinct modules: the Spectral Edge Operator (SEO) and the Spectral Saliency Generator (SSG). The bottleneck integrates a Mixed-Frequency Attention Module, featuring frequency-dependent attention heads. The deep edge feature is denoted by $\boldsymbol{F}_{DE}$, whereas $\boldsymbol{G}_1$ and $\boldsymbol{G}_2$ denote split saliency features. (b) SEO detects edge information by calculating the spectral angular distance (SAD) result's gradient. (c) SSG generates saliency maps $\boldsymbol{I}_S^1$, $\boldsymbol{I}_S^2$, and $\boldsymbol{I}_S^3$ by estimating the difference between pyramid levels. (d) The decoder preserves low-level encoder features $\boldsymbol{R}_1$ and $\boldsymbol{R}_2$, and the shallow edge feature $\boldsymbol{F}_{SE}$ to generate better saliency maps. (e) Edge Detection Module (EDM) generates an edge feature.}
        \label{overall}
\end{figure*}

\subsection{Hyperspectral Salient Object Detection}
Despite the significant advancements in SOD, the field of hyperspectral imaging remains relatively new in this context. Wilson \etal~\cite{wilson1997perceptual} introduced the concept of contrast sensitivity saliency to fuse different bands and visualize hyperspectral remote sensing images. Subsequently, dimension reduction techniques and Itti's attention model~\cite{itti1998model} were incorporated into HSOD. Moan \etal~\cite{moan2011saliency} divided the spectrum of a hyperspectral image into three regions and employed principal component analysis (PCA) to extract the first principal component of each region. Zhang \etal~\cite{zhang2008hyperspectral} utilized both real color and PCA images for visualization. However, although these methods offer computational efficiency, they suffer from inevitable information loss due to feature reduction.

Recently, Imamoglu \etal~\cite{imamoglu2018hyperspectral, imamouglu2019salient} introduced the first dataset specifically designed for HSOD. They employed a manifold ranking algorithm and extracted features using a self-supervised CNN to generate saliency maps. Similarly, Huang \etal~\cite{huang2021salient} utilized a CNN with two channels to extract spatial and spectral features separately, which were subsequently fused to optimize the saliency values of both foreground and background cues, leading to improved detection performance but at the expense of high computational complexity as well as low computing speed.
Our model takes Spectral Saliency and Spectral Edge as high-level inputs. Instead of employing clustering algorithms, it employs an end-to-end neural network for salient object detection, effectively addressing the abovementioned drawbacks.

\subsection{Attention Mechanism in Salient Object Detection}
Although the attention mechanism is crucial for SOD, it was first employed in image classification~\cite{Mnih_2014_Recurrent}. Yin \etal~\cite{Yin_2016_ABCNN} were the first attempt to incorporate the attention mechanism into CNNs. The subsequent emergence of the self-attention mechanism demonstrated a powerful ability to capture features with long-range dependencies, witnessing great success in machine translation~\cite{vaswani2017attention} and image classification~\cite{dosovitskiy2020image}, \textit{etc}.

The potential of the self-attention mechanism in SOD was first recognized by Liu \etal~\cite{liu2021visual} and Zhang \etal~\cite{zhang2021learning}. Subsequent studies~\cite{zhudftr,liu2021swinnet,yang2022bi} have further expanded the application of self-attention in SOD tasks. 
In our work, we introduce a novel attention mechanism called Mixed-frequency Attention, which employs one attention head to concentrate solely on saliency information while another focuses on the interaction between edge and saliency information. This pioneering approach represents the first integration of the attention mechanism into the HSOD task.

\section{Method}
\label{sec:method}
Given an HSI denoted as $\boldsymbol{I} \in \mathbb{R}^{\mathrm{H} \times \mathrm{W} \times \mathrm{C}}$, the primary objective of hyperspectral salient object detection is to generate a saliency map denoted as $\boldsymbol{S} \in \mathbb{R}^{\mathrm{H} \times \mathrm{W} \times \mathrm{1}}$, which provides information about the location of the salient object within the HSI. Such a mapping process can be formulated as: 
\begin{equation}
	\boldsymbol{S} = \boldsymbol{\Phi}(\boldsymbol{I}).
\end{equation}
The mapping function $\boldsymbol{\Phi}(\cdot)$ is implemented by a novel Spectrum-driven Mixed-frequency Network (SMN).

\cref{overall} (a) presents the comprehensive architecture of SMN, which encompasses four key steps: Spectral Saliency and Spectral Edge extraction, frequency-specified embeddings, Mixed-frequency Attention, and saliency-edge-aware decoding.
To elaborate, the first step involves the extraction of Spectral Saliency and Spectral Edge images using dedicated plug-and-play operators. These images are subsequently incorporated into deep saliency or edge features by means of frequency-specified embeddings.
Moving on to the third step, a Mixed-frequency Attention module is employed to fully harness the complementary nature of these features. This enables the generation of a mixed-frequency feature that encompasses comprehensive edge and saliency information.
Lastly, in the fourth step, the saliency-edge-aware decoder gradually upscales the mixed-frequency feature while simultaneously preserving the fine-grained edge details and shallow saliency information. Ultimately, this decoding process culminates in the production of the final saliency map.

\subsection{Spectral Saliency Generator}
\label{SSG}
The Spectral Saliency Generator (SSG) is a stand-alone layer responsible for generating Spectral Saliency maps. As illustrated in~\cref{overall} (c), these maps are produced by computing the ``center-surround" similarity between pairs of Gaussian pyramid layers, constructed from the input HSI. The Spectral Saliency maps provide an approximate indication of the salient object's location and serve as the low-frequency input to SMN.

Specifically, the input HSI undergoes an initial downsampling process using Gaussian downsampling operations. This process involves applying depth-wise convolution with a fixed Gaussian weight to create a Gaussian pyramid with $N$ layers ($N=8$). Through Gaussian downsampling, the spatial dimensions of the image decrease as the scale increases, and each pixel's information is influenced by a larger neighborhood of pixels. This enables the assessment of the saliency value between a ``center" pixel at point $(i,j)$ and its ``surround" pixel. The comparison is executed via the calculation of the spectral angular distance (SAD) between spectral vectors \( \boldsymbol{v}_c \) and \( \boldsymbol{v}_s \), which are derived from the \( c \)-th and \( s \)-th layers of the Gaussian pyramid, respectively. The value of the saliency map $\boldsymbol{I}_{S}$ at this point is computed as follows:
\begin{equation}
    \label{SAD}
    \boldsymbol{I}_{S}(i,j) = \text{arccos} \left(\frac{\boldsymbol{v}_c^T(i,j) \boldsymbol{v}_s(i,j)}{\| \boldsymbol{v}_c(i,j) \| \| \boldsymbol{v}_s(i,j) \|}\right),
\end{equation}
\pexp $\left\Vert \cdot \right\Vert$ is the Euclidean norm of a vector. In this context, the layer index $c$ of the ``center" pixel takes on values from the set $\{ 2,3,4 \}$, and $s$ is determined as $c+3$. By performing the aforementioned calculation for each point of the image, the saliency map of the entire HSI can be obtained. Three values of the layer index $c$ yield different saliency maps, denoted as $\{\boldsymbol{I}_{S}^k \}^{3}_{k=1}$. The dimensions of each saliency map are \( \mathrm{H} \times \mathrm{W} \times 1 \).

\subsection{Spectral Edge Operator} \label{SEO}
The blurring of object edges and loss of high-frequency information in Spectral Saliency images are consequences of Gaussian downsampling. Employing such images for saliency detection could result in less sharp or even erroneous edges in the detection results. To incorporate high-frequency details into the SMN, we have devised a module called the Spectral Edge Operator (SEO). Drawing inspiration from edge detection operators like the Canny operator, SEO extracts Spectral Edge images by computing the gradient of the SAD in the vicinity of each pixel.

Specifically, for a point $(i, j)$ on the HSI, assume its neighborhood size is $\mathrm{H}^\prime \times \mathrm{W}^\prime$. The SAD between this point and any point $(p, q)$ within its neighborhood can be computed, resulting in a local spectral similarity map $\boldsymbol{M} \in \mathbb{R}^{\mathrm{H}^\prime \times \mathrm{W}^\prime}$. This process can be formulated as follows:
\begin{equation}
    \boldsymbol{M}(p,q) = \text{arccos} \left(\frac{\boldsymbol{v}(i,j)^T \boldsymbol{v}(p,q)}{\left\Vert\boldsymbol{v}(i,j)\right\Vert \left\Vert\boldsymbol{v}(p,q)\right\Vert}\right),
\end{equation}
\pexp $\boldsymbol{v}(i,j)$ and $\boldsymbol{v}(p,q)$ represent the spectral vectors of points at $(i,j)$ and $(p,q)$, respectively. To compute the value of the edge image $\boldsymbol{I}_E$ at $(i, j)$, gradient convolution kernels the same size as the neighborhood, denoted as $\boldsymbol{G}_x$ and $\boldsymbol{G}_y$, are applied to the local spectral similarity map $\boldsymbol{M}$:
\begin{equation}
    \label{edge}
    \boldsymbol{I}_{E}(i,j) = \left |\boldsymbol{G}_x \ast \boldsymbol{M} \right | + \left |\boldsymbol{G}_y \ast \boldsymbol{M} \right |.
\end{equation}
In \cref{overall} (b), the specific content of $\boldsymbol{G}_x$ is depicted, with a size of 5. $\boldsymbol{G}_y$ is the transpose of $\boldsymbol{G}_x$. By applying these operations to every pixel in the image, the Spectral Edge image for the entire image can be obtained. Efficient CUDA kernels are employed to expedite the computation. Three kernels of varying sizes are employed to extract Spectral Edge images $\{ \boldsymbol{I}_{E}^k \}_{k=1}^3$, where the dimensions of each image are \( \mathrm{H} \times \mathrm{W} \times 1 \).

\subsection{Frequency-specified Embeddings}
Both the Spectral Saliency Generator and the Spectral Edge Operator produce three sets of Spectral Saliency and Spectral Edge maps, respectively. These maps capture distinct and important information for hyperspectral salient object detection. To leverage this valuable information, the maps are transformed into deep saliency or edge features using frequency-specified embeddings.

\nbf{Deep Saliency Feature}
The deep saliency feature is obtained through a low-frequency embedding process. To generate this feature, the Spectral Saliency images, denoted as $\boldsymbol{I}_{S}^{1}$, $\boldsymbol{I}_{S}^{2}$, and $\boldsymbol{I}_{S}^{3}$, are concatenated along the channel dimension:
\begin{equation}
    \boldsymbol{F}_S = [\boldsymbol{I}_{S}^{1}, \boldsymbol{I}_{S}^{2}, \boldsymbol{I}_{S}^{3}], 
\end{equation}
\pexp $\boldsymbol{F}_S$ is the concatenation result. By concatenating the Spectral Saliency images, rather than sequentially feeding them into the network, the computational complexity is reduced, leading to improved inference speed. $\boldsymbol{F}_S$ are then transformed into a deep saliency feature $\boldsymbol{F}_{DS}$ as:
\begin{equation}
    \boldsymbol{F}_{DS} = \boldsymbol{f}_\text{L}(\boldsymbol{F}_S),
\end{equation}
where $\boldsymbol{f}_\text{L}(\cdot)$ represents the low-frequency embedding process.

Let $\boldsymbol{R} = \{\boldsymbol{R}_i|i=1,2,3\}$ represent the multi-stage features obtained from the low-frequency embedding process, where each stage captures $\frac{1}{4}$, $\frac{1}{8}$, and $\frac{1}{16}$ of the input feature, respectively. The deep saliency feature refers to the last stage of the low-frequency embedding: $\boldsymbol{F}_{DS} = \boldsymbol{R}_3 \in \mathbb{R}^{\frac{\mathrm{H}}{16} \times \frac{\mathrm{W}}{16} \times \mathrm{C}_S}$, where $\mathrm{C}_S$ is the channel dimension of the deep saliency feature.
Furthermore, the shallow saliency information captured in $\boldsymbol{R}_1$ and $\boldsymbol{R}_2$ is retained for later use in the decoder stage.

\nbf{Deep Edge Feature}
A deep edge feature can be obtained through high-frequency embedding, which incorporates a downsampling block and an Edge Detection Module (EDM)~\cite{li2022multi}. The high-frequency embedding takes concatenated Spectral Edge images as input and generates a shallow edge feature $\boldsymbol{F}_{SE}$, which can be expressed as follows:
\begin{equation}
    \boldsymbol{F}_{SE} = \boldsymbol{f}_\text{D}(\left [ \boldsymbol{I}_{E}^{1}, \boldsymbol{I}_{E}^{2}, \boldsymbol{I}_{E}^{3} \right ]),
\end{equation}
where $\boldsymbol{f}_\text{D}(\cdot)$ denotes the downsampling block implemented using two $\textit{conv}3\times3$ layers with a stride of 2, followed by batch normalization layers and ReLU activation functions. The shallow edge feature $\boldsymbol{F}_{SE}$ is then transformed into a deep edge feature $\boldsymbol{F}_{DE}$ using EDM $\boldsymbol{f}_\text{E}(\cdot)$:
\begin{equation}
    \boldsymbol{F}_{DE} = \boldsymbol{f}_\text{E}(\boldsymbol{F}_{SE}).
\end{equation}
Here, the shape of $\boldsymbol{F}_{DE}$ is ${\frac{\mathrm{H}}{16} \times \frac{\mathrm{W}}{16} \times \mathrm{C}_E}$, with $\mathrm{C}_E$ denoting the dimension of the edge feature. The specific composition of EDM is illustrated in~\cref{overall} (e).

To enhance the quality of the generated edge features, we transform $\boldsymbol{F}_{DE}$ into an edge image $\boldsymbol{M}_E$, which is constrained by a ground truth edge image. The generation process of $\boldsymbol{M}_E$ can be mathematically described as follows:
\begin{equation}
    \boldsymbol{M}_E = \boldsymbol{f}_\text{CU}(\boldsymbol{F}_{DE}),
\end{equation}
where $\boldsymbol{f}_\text{CU}(\cdot)$ is implemented using a $\textit{conv}1\times1$ layer followed by an upsampling layer. The resulting edge map $\boldsymbol{M}_E$ is a single-channel grayscale image with the same dimensions as the original image. The ground truth edge image $\boldsymbol{E}$ is generated using an edge detector~\cite{su2021pdc}, which is obtained by combining two edge maps:
\begin{equation}
    \label{gt_edge}
    \boldsymbol{E}=\boldsymbol{e}(\boldsymbol{I}_{FC})+\boldsymbol{e}(\boldsymbol{I}_{S}^\prime),
\end{equation}
where $\boldsymbol{e}(\cdot)$ represents the edge detector. $\boldsymbol{I}_{FC}$ corresponds to the false-color image rendered from HSI, while $\boldsymbol{I}_{S}^\prime$ is the sum of the previously generated spectral saliency maps $\boldsymbol{I}_{S}$.

\subsection{Mixed-frequency Attention}
The Mixed-frequency Attention (MA) module facilitates the comprehensive interaction and fusion of deep features with different frequencies. It encompasses two essential components: the high-frequency head and the low-frequency head.
In the high-frequency head, a cross-attention mechanism is employed to capture the relationship between high-frequency edge features and low-frequency saliency features. This interaction enables the refinement of saliency representations under the constraints imposed by edge information.
Conversely, in the low-frequency head, a self-attention mechanism is applied to the saliency feature itself. This enables the generation of more precise and accurate saliency representations by emphasizing relevant saliency information within the feature.

\nbf{Saliency Feature Division} 
The saliency feature, denoted by \( \boldsymbol{F}_{DS} \) and having a channel dimension \( \mathrm{C}_S \), is uniformly partitioned into two groups along the channel dimension:
\begin{equation}
    \label{saliency_heads}
    \begin{aligned}
        \boldsymbol{G}_1 &= \boldsymbol{F}_{DS} \left (:, :, 0:\lfloor \frac{\mathrm{C}_S}{2} \rfloor \right ), \\
        \boldsymbol{G}_2 &= \boldsymbol{F}_{DS} \left (:, :, \lfloor \frac{\mathrm{C}_S}{2} \rfloor:\mathrm{C}_S \right ),
    \end{aligned}
\end{equation}
where $\boldsymbol{G}_1$ and $\boldsymbol{G}_2$ represent the resulting groups after the division. Each group has dimensions of $\frac{\mathrm{H}}{16} \times \frac{\mathrm{W}}{16} \times \lfloor \frac{\mathrm{C}_S}{2} \rfloor$. These groups are subsequently fed into different heads of the Mixed-frequency Attention module for further processing.

\nbf{High-frequency Attention Head}
The high-frequency attention head incorporates both the deep saliency and edge features, enabling their interaction to enhance the accuracy of saliency detection. 
We employ a neighborhood attention mechanism (NAM)~\cite{Hassani_2023_NAT} to confine the receptive field of the \textit{query} to its local neighborhood, enhancing its sensitivity to edge information while reducing computational complexity. Suppose the input matrices of the NAM are denoted as $\boldsymbol{X}$ and $\boldsymbol{Y}$, respectively. The NAM can be defined as:
\begin{equation}
\label{neighborhood attention}
    \boldsymbol{f}_\text{NAM}(\boldsymbol{X}, \boldsymbol{Y}) = \sigma \left(\boldsymbol{Q}_{i,j} \boldsymbol{K}^T_{\rho(i,j)}+B_{i,j}\right) \boldsymbol{V}_{\rho(i,j)},
\end{equation}
where $\rho(i,j)$ represents the neighborhood of a pixel at position $(i,j)$, and $B_{i,j}$ denotes the relative positional bias. The function $\sigma(\cdot)$ corresponds to the Sigmoid function. The \textit{query} matrix $\boldsymbol{Q}$ is derived from $\boldsymbol{X}$, while the \textit{key} matrix $\boldsymbol{K}$ and the \textit{value} matrix $\boldsymbol{V}$ are obtained from $\boldsymbol{Y}$:
\begin{equation}
\begin{aligned}
    \label{cross_QKV}
    \boldsymbol{Q} = \boldsymbol{X}\boldsymbol{W}^{Q}, \boldsymbol{K} = \boldsymbol{Y}\boldsymbol{W}^{K},\boldsymbol{V} = \boldsymbol{Y}\boldsymbol{W}^{V},
\end{aligned}
\end{equation}
where $\boldsymbol{W}^Q, \boldsymbol{W}^K,\boldsymbol{W}^V$ represent learnable parameters implemented through linear projection.

The cross-attention between the deep edge feature $\boldsymbol{F}_{DE}$ and the first group of deep saliency feature $\boldsymbol{G}_{1}$ is computed:
\begin{equation}
\label{cross_attn}
    \boldsymbol{F}_{H} = \boldsymbol{f}_\text{NAM}(\boldsymbol{F}_{DE}, \boldsymbol{G}_1),
\end{equation}
\pexp $\boldsymbol{F}_{H}$ represents the refined saliency feature, constrained by edge information.

\nbf{Low-frequency Attention Head} 
The low-frequency attention heads exclusively receive deep saliency features, utilizing a self-attention mechanism to capture more precise representations of salient objects. 
To alleviate computational complexity, the NAM is employed. The self-attention operation is employed for the low-frequency attention result $\boldsymbol{F}_{L}$:
\begin{equation}
\label{self_attn}
    \boldsymbol{F}_{L} = \boldsymbol{f}_\text{NAM}(\boldsymbol{G}_2, \boldsymbol{G}_2).
\end{equation}
After the low-frequency attention, $\boldsymbol{F}_{L}$ serves as a more accurate saliency representation in comparison to the input $\boldsymbol{G}_2$. It should be noted that the kernel size of the NAM differs between the two heads, reflecting the differences in input and objectives for the different frequency heads.

\nbf{Frequency Convergence} 
To integrate the frequency-specific features, the outputs from the high-frequency and low-frequency attention heads are concatenated along the feature dimension, obtaining a mixed-frequency feature $\boldsymbol{F}_\text{out}$ that combines comprehensive and effectively integrated edge and saliency information:
\begin{equation}
    \begin{aligned}
    \boldsymbol{F}_\text{out} = \delta(\boldsymbol{f}_\text{C}([\boldsymbol{F}_{H},\boldsymbol{F}_{L}])),
\end{aligned}
\end{equation}
where $\delta(\cdot)$ is implemented by the ReLU activation function, and $\boldsymbol{F}_{C}$ represents the $conv1\times1$ operation.

\subsection{Saliency-edge-aware Decoder}
As illustrated in~\cref{overall} (d), the saliency-edge-aware decoder employs a cascading structure of convolutional layers, allowing for the gradual upscaling of the mixed-frequency feature. This process ensures the fusion of shallow features from the encoder, preserving intricate details and saliency information.

The cascaded decoder architecture consists of five convolutional layers, each accompanied by a batch normalization layer, a ReLU activation function, and an interpolation operation. Let $\boldsymbol{D}_\text{in} \in \{ \boldsymbol{D}_\text{in}^i|i=1,2,3,4,5\}$ and $\boldsymbol{D}_\text{out} \in \{ \boldsymbol{D}_\text{out}^i|i=1,2,3,4,5\}$ represent the input and output of these convolutional layers, respectively. To preserve shallow information, the shallow edge information $\boldsymbol{F}_{SE}$ is concatenated with $\boldsymbol{D}_\text{out}^2$ along the channel dimension:
\begin{equation}
    \boldsymbol{D}_\text{in}^3 = \left [\boldsymbol{D}_\text{out}^2, \boldsymbol{F}_{SE} \right ].
\end{equation}
The saliency information $\boldsymbol{R}_1$ and $\boldsymbol{R}_2$ are separately upsampled to match the spatial dimension, and the resulting outputs are concatenated in the channel dimension and added to $\boldsymbol{D}_\text{out}^3$:
\begin{equation}
    \boldsymbol{D}_\text{in}^4 = \boldsymbol{D}_\text{out}^3 + \left [ \boldsymbol{f}_\text{tc1}(\boldsymbol{R}_1), \boldsymbol{f}_\text{tc2}(\boldsymbol{R}_2) \right ], 
\end{equation}
where $\boldsymbol{f}_\text{tc1}(\cdot)$ and $\boldsymbol{f}_\text{tc2}(\cdot)$ represent transposed convolutional layers. By incorporating edge information through concatenation and saliency information through summation, we reduce the modification of shallow information, resulting in a less complex network that retains more shallow information.

\subsection{Hybrid Loss Function}
During the training process, certain intermediate results are supervised to ensure the precise extraction of saliency or edge features. To achieve this, a hybrid loss function $\mathcal{L}$ is utilized, given by the equation:
\begin{equation}
	\mathcal{L}=\mathcal{L}_\text{edge}+\mathcal{L}_\text{final},
\end{equation}
\pexp \( \mathcal{L}_\text{edge} \) and \( \mathcal{L}_\text{final} \) represent the loss associated with edge detection and the final saliency map, respectively. Further details will be provided subsequently.

\nbf{Binary Cross-entropy Loss}
The binary cross-entropy (BCE) loss function is defined as:
\begin{equation}
    \label{eq: equation BCE}
    \mathcal{L}_{\text{BCE}}(\boldsymbol{X},\boldsymbol{Y})=-\sum {\left [ \boldsymbol{X} \text{log}(\boldsymbol{Y})+(1-\boldsymbol{X}) \text{log}(1-\boldsymbol{Y}) \right ]},
\end{equation}
where $\boldsymbol{X}$ represents the ground-truth values and $\boldsymbol{Y}$ corresponds to the input matrix. In the case of edge map $\boldsymbol{M}_E$, it is supervised using the BCE loss with the edge ground truth $\boldsymbol E$ as follows:
\begin{equation}
        \mathcal{L}_\text{edge} = \mathcal{L}_{\text{BCE}}(\boldsymbol{M}_E,\boldsymbol{E}).
\end{equation}

\nbf{Intersection Over Union Loss}
In accordance with Qin \etal~\cite{Qin_2019_BASNet}, we incorporate the intersection over union (IoU) loss function, defined as:
\begin{equation}
\footnotesize
    \mathcal{L}_{\text{IoU}}(\boldsymbol{X}, \boldsymbol{Y}) = 1 - \frac{\sum\limits_{r=1}^{\mathrm{H}}\sum\limits_{c=1}^{\mathrm{W}} \boldsymbol{X}(r,c)\boldsymbol{Y}(r,c)}{\sum\limits_{r=1}^{\mathrm{H}}\sum\limits_{c=1}^{\mathrm{W}} [\boldsymbol{X}(r,c)+\boldsymbol{Y}(r,c)-\boldsymbol{X}(r,c)\boldsymbol{Y}(r,c)]},
\end{equation}
where $\boldsymbol{X}$ and $\boldsymbol{Y}$ denote the input and ground-truth matrices, respectively, and $\mathrm{H}$ and $\mathrm{W}$ represent the height and width. This loss function is employed to supervise the final saliency map $\boldsymbol{S}$:
\begin{equation}
    \mathcal{L}_\text{final} = \mathcal{L}_{\text{IoU}}(\boldsymbol{S},\boldsymbol{G}) + \mathcal{L}_{\text{BCE}}(\boldsymbol{S},\boldsymbol{G}),
\end{equation}
where $\boldsymbol{G}$ denotes the ground-truth saliency map.

\section{Experiments}
\label{sec:experiments}

\subsection{Experimental Settings}
\label{exp settings}
\nbf{Datasets} 
Two datasets are utilized for assessing the performance of SMN: HS-SOD~\cite{imamoglu2018hyperspectral} and our dataset, HSOD-BIT. HS-SOD comprises 60 HSIs with a spectral range of 380-780$\rm{nm}$ at intervals of 5$\rm{nm}$, and a spatial resolution of $768\times1024$ pixels. For the purpose of evaluation, 48 HSIs are allocated for training, while 12 HSIs are reserved for testing. On the other hand, HSOD-BIT encompasses 319 HSIs, each possessing a spatial resolution of $1240\times1680$ pixels and a spectral range of 400-1000$\rm{nm}$ with intervals of 3$\rm{nm}$. In HSOD-BIT, 255 HSIs are employed for training, while 64 HSIs are utilized for testing. Both datasets comprise RGB images as well as binarized ground-truth images that correspond to the respective HSIs.

\nbf{Evaluation Metrics}
The assessment of saliency map detection performance necessitates the utilization of established evaluation metrics. The metrics are delineated as follows:

Mean absolute error (MAE) quantifies the pixel-level discrepancy between the saliency map $\boldsymbol{S}$ and the ground truth image $\boldsymbol{G}$:
\begin{equation}
    \text{MAE}=\frac{1}{\mathrm{W} \times \mathrm{H}} \sum_{x=1}^\mathrm{W} \sum_{y=1}^\mathrm{H} | \boldsymbol{S}_{xy}-\boldsymbol{G}_{xy} |,
\end{equation}
\pexp $\mathrm{W}$ and $\mathrm{H}$ denote the width and height of the input image, respectively.

S-measure~\cite{SMeasure} ($S_\alpha$) evaluates the structural fidelity of the saliency map $\boldsymbol{S}$ and is defined as a weighted sum of region similarity $S_r$ and object similarity $S_o$:
\begin{equation}
    S_\alpha = \alpha \times S_r(\boldsymbol{S}, \boldsymbol{G}) + (1 - \alpha) \times S_o(\boldsymbol{S}, \boldsymbol{G}).
\end{equation}
For the definitions of $S_r$ and $S_o$, the reader is referred to \cite{SMeasure}. We adopt $\alpha=0.5$, as recommended in \cite{SMeasure}.

Precision-Recall (PR) curve~\cite{borji_2015_salient} serves as a conventional metric for saliency evaluation. It is derived by thresholding the saliency map from 0 to 255 and subsequently computing precision and recall at each threshold level:
\begin{equation}
    \text { Precision }=\frac{|\boldsymbol{B} \cap \boldsymbol{G}|}{|\boldsymbol{B}|}, \text { Recall }=\frac{|\boldsymbol{B} \cap \boldsymbol{G}|}{|\boldsymbol{G}|},
\end{equation}
\pexp $\boldsymbol{B}$ and $\boldsymbol{G}$ denote the binarized saliency maps and the ground truth, respectively.

F-measure~\cite{F-measure} ($F_\beta$) is the harmonic mean of precision and recall, formulated as:
\begin{equation}
    F_\beta = \frac{(1 + \beta^2) \times \text{Precision} \times \text{Recall}}{\beta^2 \times \text{Precision} + \text{Recall}}.
\end{equation}
We employ the maximum F-measure, denoted as $F_\beta^\text{max}$, for comparative analyses. The value of $\beta^2$ is set to 0.3, as suggested in \cite{F-measure}.

Receiver Operating Characteristic (ROC) curve~\cite{borji_2015_salient} is characterized by the true positive rate (TPR) and false positive rate (FPR):
\begin{equation}
    \text { TPR }=\frac{|\boldsymbol{B} \cap \boldsymbol{G}|}{|\boldsymbol{G}|}, \text { FPR }=\frac{|\boldsymbol{B} \cap \bar{\boldsymbol{G}}|}{|\bar{\boldsymbol{G}}|}.
\end{equation}
Here, $\bar{\boldsymbol{G}}$ denotes the complement of the ground truth $\boldsymbol{G}$. Area Under Curve (AUC) is the total area under the ROC curve.

Correlation Coefficient (CC) ~\cite{bylinskii_2019_what} measures the statistical correlation between the saliency map $\boldsymbol{S}$ and ground truth $\boldsymbol{G}$:
\begin{equation}
    \text{CC} = \frac{\sigma(\boldsymbol{S}, \boldsymbol{G})}{\sigma(\boldsymbol{S}) \times \sigma(\boldsymbol{G})},
\end{equation}
\pexp  $\sigma(\boldsymbol{S}, \boldsymbol{G})$ is the covariance between $\boldsymbol{S}$ and $\boldsymbol{G}$. Overall, a better HSOD saliency detector shall have a smaller MAE and larger other metrics.

\begin{table*}[htp]
\centering
\caption{Quantitative results on HSOD-BIT and HS-SOD datasets. `-R': ResNet18~\cite{ResNet}, `-S': Swin-tiny~\cite{liu2021swin}, `-P': PVTv2-b1~\cite{wang2021pvtv2}.}
\label{Dingliang}
\setlength{\tabcolsep}{3.5mm}{
\begin{tabular}{l|ccccc|ccccc} 
\toprule
Datasets & \multicolumn{5}{c|}{HSOD-BIT}                 & \multicolumn{5}{c}{HS-SOD}                     \\ 
\midrule
Metrics  & MAE $\downarrow$  & $S_\alpha$ $\uparrow$ & $F_\beta^\text{max}$ $\uparrow$ & AUC $\uparrow$  & CC  $\uparrow$  & MAE $\downarrow$  & $S_\alpha$ $\uparrow$ & $F_\beta^\text{max}$ $\uparrow$ & AUC $\uparrow$  & CC  $\uparrow$   \\ 
\midrule
Itti~\cite{itti1998model}    & 0.247 & 0.532     & 0.374     & 0.801 & 0.355 & 0.257 & 0.488     & 0.271     & 0.783 & 0.225  \\
SAD~\cite{liang2013salient}     & 0.203 & 0.552     & 0.390     & 0.830 & 0.397 & 0.203 & 0.500     & 0.244     & 0.778 & 0.223  \\
SED~\cite{liang2013salient}     & 0.130 & 0.500     & 0.343     & 0.753 & 0.303 & 0.132 & 0.470     & 0.291     & 0.793 & 0.201  \\
SG~\cite{liang2013salient}      & 0.182 & 0.543     & 0.338     & 0.791 & 0.370 & 0.196 & 0.530     & 0.274     & 0.808 & 0.268  \\
SUDF~\cite{imamouglu2019salient}    & 0.150 & 0.685     & 0.544     & 0.918 & 0.671 & 0.242 & 0.498     & 0.275     & 0.723 & 0.250  \\ 
BASNet~\cite{Qin_2019_BASNet}  & 0.040 & 0.849     & 0.779     & 0.919 & 0.785 & 0.071 & 0.743     & 0.605     & 0.843 & 0.625  \\
U2Net~\cite{Qin_2020_U2Net}   & 0.034 & 0.870     & 0.829     & 0.942 & 0.830 & 0.076 & 0.734     & 0.617     & 0.854 & 0.631  \\ 
\midrule
SMN-R (Ours)     & 0.039 & 0.869     & 0.854     & 0.969 & 0.849 & 0.069 & 0.767     & 0.682     & 0.903 & 0.684  \\
SMN-S (Ours)     & \textbf{0.032} & 0.891     & 0.868     & 0.971 & 0.870 & 0.079 & 0.737     & 0.659     & 0.899 & 0.635  \\
SMN-P (Ours)     & 0.034 & \textbf{0.892}     & \textbf{0.872}     & \textbf{0.981} & \textbf{0.874} & \textbf{0.068} & \textbf{0.788}     & \textbf{0.723}     & \textbf{0.916} & \textbf{0.718}  \\
\bottomrule
\end{tabular}}
\end{table*}

\nbf{Implementation Details} 
For the purpose of reducing memory cost and computational complexity, we performed downsampling on the original HSIs both spatially and spectrally. As a result, the HSIs were transformed into a spatial resolution of $224 \times 224$ pixels and consisted of 50 spectral channels. To augment the data, we employed horizontal flip and random crop techniques.
In the low-frequency embedding phase, ResNet18~\cite{ResNet}, Swin-tiny~\cite{liu2021swin}, and PVTv2-b1~\cite{wang2021pvtv2} were utilized as base architectures, initialized with weights pre-trained on the ImageNet1k dataset. The models are denoted as SMN-R, SMN-S, and SMN-P, respectively. To implement cross-attention, we modified the neighborhood attention mechanism accordingly. The kernel size for the high-frequency and low-frequency attention heads was set to 13 and 9, respectively.
Our model was trained on a single NVIDIA RTX 3090 GPU with an Intel XEON Gold 5218R CPU. Stochastic gradient descent (SGD) with a momentum optimizer was employed for training, spanning a total of 100 epochs. A warm-up and linear decay strategy was employed to calibrate the maximum learning rate to \(2 \times 10^{-2}\) (for Swin-tiny and PVTv2-b1, it was set to \(7 \times 10^{-3}\)). The batch size was configured to 5.

\nbf{Competing methods} 
Itti's model \cite{itti1998model} serves as the baseline model for HSOD. Initially, we compare our model with several conventional methods proposed by Liang \etal\cite{liang2013salient}, namely spectral angular distance (SAD), spectral Euclidean distance (SED), and spectral grouping (SG). 
In order to compare with open-source state-of-the-art methods, we also include SUDF proposed by Imamouglu \etal~\cite{imamouglu2019salient} in the comparison. For the sake of fairness, SUDF retains the default parameter settings. Furthermore, we compare our SMN with two classical RGB-image-based SOD methods, BASNet~\cite{Qin_2019_BASNet} and U2Net~\cite{Qin_2020_U2Net}, to validate the necessity of developing HSOD methods.

\subsection{Results on HSOD-BIT}
\nbf{Quantitative Results}
The quantitative comparison results on HSOD-BIT can be found in \cref{Dingliang}. Regardless of the backbone network employed, our SMN consistently outperforms both traditional methods and SUDF. Specifically, our SMN-R achieves impressive scores of $0.039$ for MAE, $0.869$ for $S_\alpha$, $0.854$ for $F_\beta^\text{max}$, $0.969$ for AUC, and $0.849$ for CC. These results surpass SUDF by $74.00\%$, $29.93\%$, $56.99\%$, $5.56\%$, and $26.53\%$, respectively. Utilizing other transformer-based backbones yields enhanced detection performance, validating the effectiveness of our SMN. Traditional methods heavily rely on manually designed low-level features and fail to effectively exploit the entire spectral information, thus limiting their ability to generate highly accurate saliency maps.

\begin{table}[!htp]
\centering
\caption{Quantitative Efficiency Analysis. `-R': ResNet18~\cite{ResNet}, `-S': Swin-tiny~\cite{liu2021swin}, `-P': PVTv2-b1~\cite{wang2021pvtv2}.}
\label{Efficiency Analysis}
\setlength{\tabcolsep}{2mm}{
\begin{tabular}{l|cccc} 
    \toprule
    Metrics  & FLOPs (G) & \#Params (M) & Speed (FPS) & $F_\beta^\text{max}$ $\uparrow$\\ 
    \midrule
    SUDF~\cite{imamouglu2019salient}    & 82.90         & \textbf{0.10}            & 0.51          & 0.544 \\ 
    BASNet~\cite{Qin_2019_BASNet}  & 127.56    & 87.06        & \textbf{51.40}      & 0.779 \\
    U2Net~\cite{Qin_2020_U2Net}   & 47.65     & 44.01        & 33.47  & 0.829 \\ 
    \midrule
    SMN-R (Ours)     & \textbf{14.58}     & 7.27         & 35.91     & 0.854  \\
    SMN-S (Ours)     & 17.23     & 16.87         & 30.17     & 0.868  \\
    SMN-P (Ours)     & 14.76     & 10.23         & 32.68     & \textbf{0.872}  \\
    \bottomrule
\end{tabular}
}
\end{table}

\begin{figure}[!htp]
    \centering
    \includegraphics[width=\linewidth]{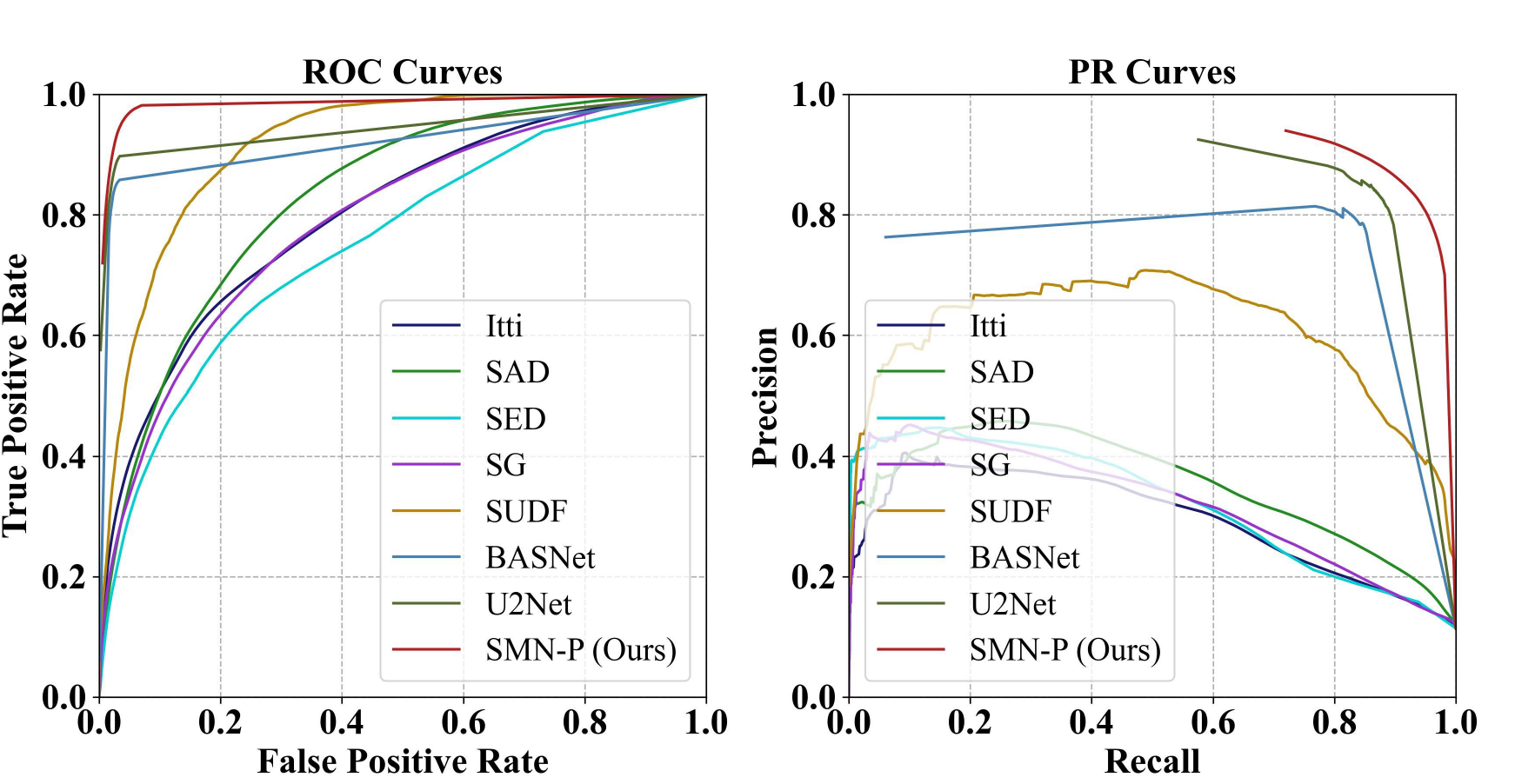}
        \caption {
        Comparison of ROC and PR curves for multiple models on our HSOD-BIT dataset. Our SMN, represented by a red line, outperforms others.
        }
    \label{PR curve on BIT}
\end{figure}

In comparison to U2Net, a classical RGB-image-based SOD method, our SMN-R exhibits superior performance in terms of $F_\beta^\text{max}$, AUC, and CC, while slightly trailing behind U2Net in the MAE and $S_\alpha$ metrics. Employing Swin-tiny and PVTv2-b1 as backbones results in detection performance substantially superior to that of U2Net. Specifically, as shown in~\cref{Dingliang}, when employing Swin-tiny as the backbone (SMN-S), we observed a $4.7\%$ increase in $F_\beta^\text{max}$ and a $3.1\%$ increase in AUC compared to U2Net. Similarly, for the PVTv2-b1 backbone (SMN-P), the $F_\beta^\text{max}$ and AUC increased by $5.2\%$ and $4.1\%$ compared to U2Net on the HSOD-BIT dataset. 

These outcomes highlight the inefficacy of employing SOD methods after simply converting HSIs into false-color images. The presence of similar colors between the foreground and background in the false-color image poses a challenge in distinguishing between them. Conversely, our SMN fully utilizes spectral information, remaining unaffected by variations in illumination conditions, thereby enabling the detection of salient objects even in challenging scenarios.

\cref{PR curve on BIT} provides a comparison of the ROC curves and PR curves between our SMN and other methods. Our SMN is denoted by the red line. Notably, the ROC curve of our SMN closely approaches the point (0, 1), while the PR curve is nearest to the point (1, 1) in comparison to the other methods. These observations indicate the superior performance of our SMN. The combined analysis of ROC curves, PR curves, and the accompanying numerical evaluation metrics serves to validate the effectiveness of our SMN.

\nbf{Efficiency Analysis} 
We conducted a comparative efficiency analysis of our SMN with other methods, including Floating Point Operations (FLOPs), number of parameters (\#Params), and inference speed (FPS). The results are shown in \cref{Efficiency Analysis}. It is worth noting that the spatial dimensions and the number of spectral channels for each method were kept at their respective default values. 
SUDF employs a CNN for feature extraction purposes only, followed by manifold learning and superpixel clustering. Consequently, it utilizes a relatively small number of parameters and exhibits a lower inference speed, at 0.1~M and 0.51~FPS, respectively. Moreover, due to the use of the entire HSI as input without spatial downsampling, SUDF incurs a high computational cost in terms of FLOPs. Our SMN demonstrates a reduction in FLOPs and an enhanced inference speed relative to SUDF. Moreover, in comparison to BASNet and U2Net, our approach significantly minimizes both the parameter and FLOPs, yet achieves commendable detection performance. 
Changing the backbone network results in a modest increase in FLOPs and the number of parameters, but does not significantly impact inference speed, while substantially enhancing detection performance. This demonstrates that SMN offers a good trade-off between computational efficiency, speed, and effectiveness.

\begin{figure*}[htp]
    \centering
    \includegraphics[width=\textwidth]{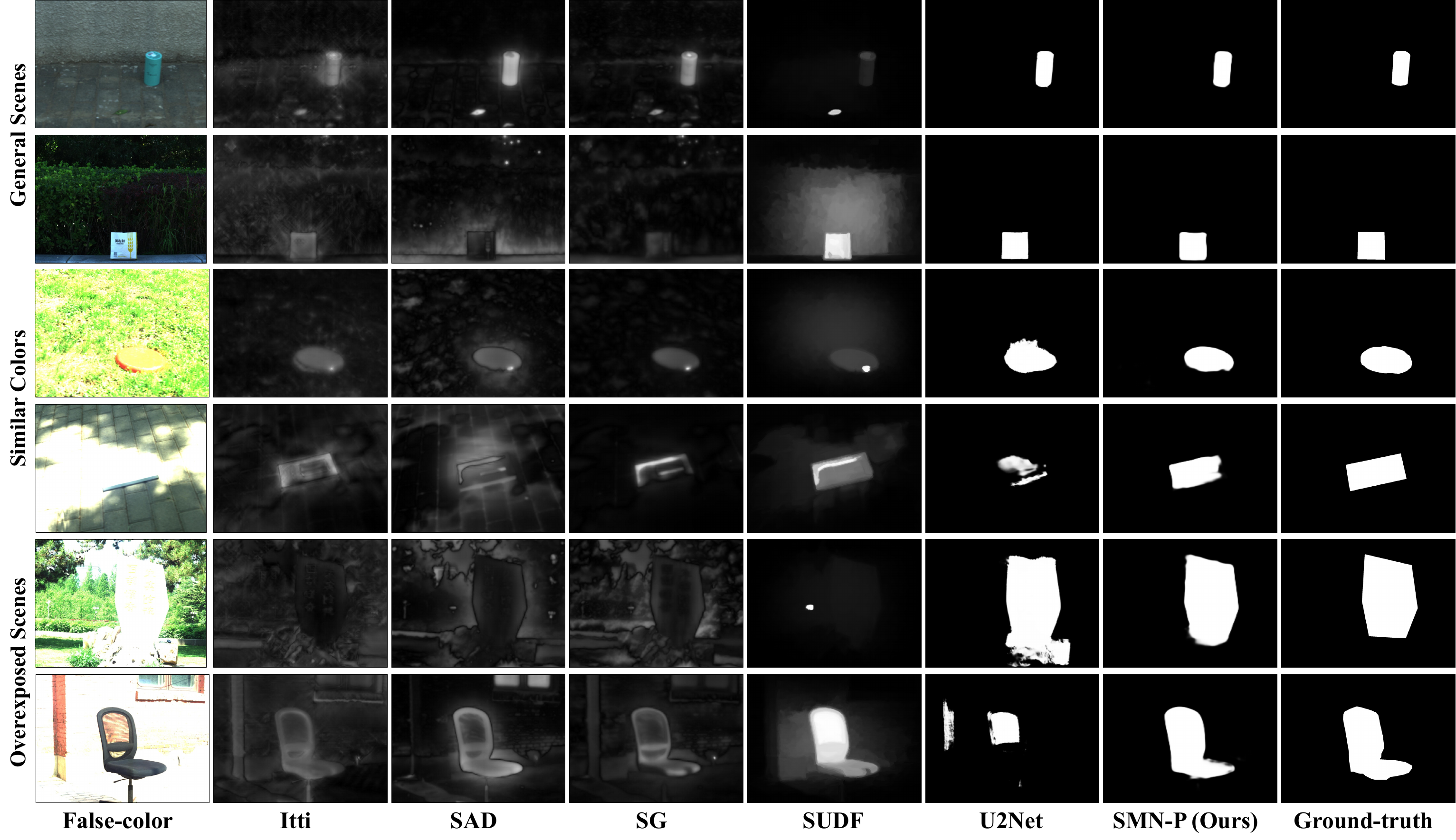}
        \caption {Qualitative results on our HSOD-BIT dataset. SMN has better detection performance in similar colors and overexposed scenes.}
    \label{qualitative}
\end{figure*}

\nbf{Qualitative Results} 
The qualitative results obtained on HSOD-BIT are presented in \cref{qualitative}. In comparison to previous HSOD approaches, our proposed SMN demonstrates the ability to accurately and comprehensively detect salient objects. For instance, in scenes on rows 1, 3, and 5, some HSOD algorithms yield misleading saliency outcomes or struggle to detect salient objects effectively.

Furthermore, we compare our SMN with a well-known RGB-image-based SOD algorithm called U2Net~\cite{Qin_2020_U2Net}. Under normal circumstances, SMN achieves comparable detection performance to U2Net: both methods produce complete detection results with sharp edges. However, in scenes where the foreground and background colors are similar, SMN exhibits more precise edge delineation compared to U2Net. Moreover, in overexposed scenes, SMN showcases higher levels of detection accuracy and completeness relative to U2Net.
This is attributed to the fact that U2Net relies solely on spatial or color information from the false-color image, rendering it ineffective in distinguishing foreground from background in scenarios where color information is lacking. Moreover, U2Net does not utilize edge information to refine saliency, leading to erroneous detection results, as illustrated in the third and fifth rows of the scenes. In contrast, our SMN approaches the problem from a spectral perspective, extracting Spectral Saliency and Spectral Edge images separately, and combining them through a specially designed Mixed-frequency Attention mechanism to leverage their complementarity.

\begin{figure}[tp]
    \centering
    \includegraphics[width=\linewidth]{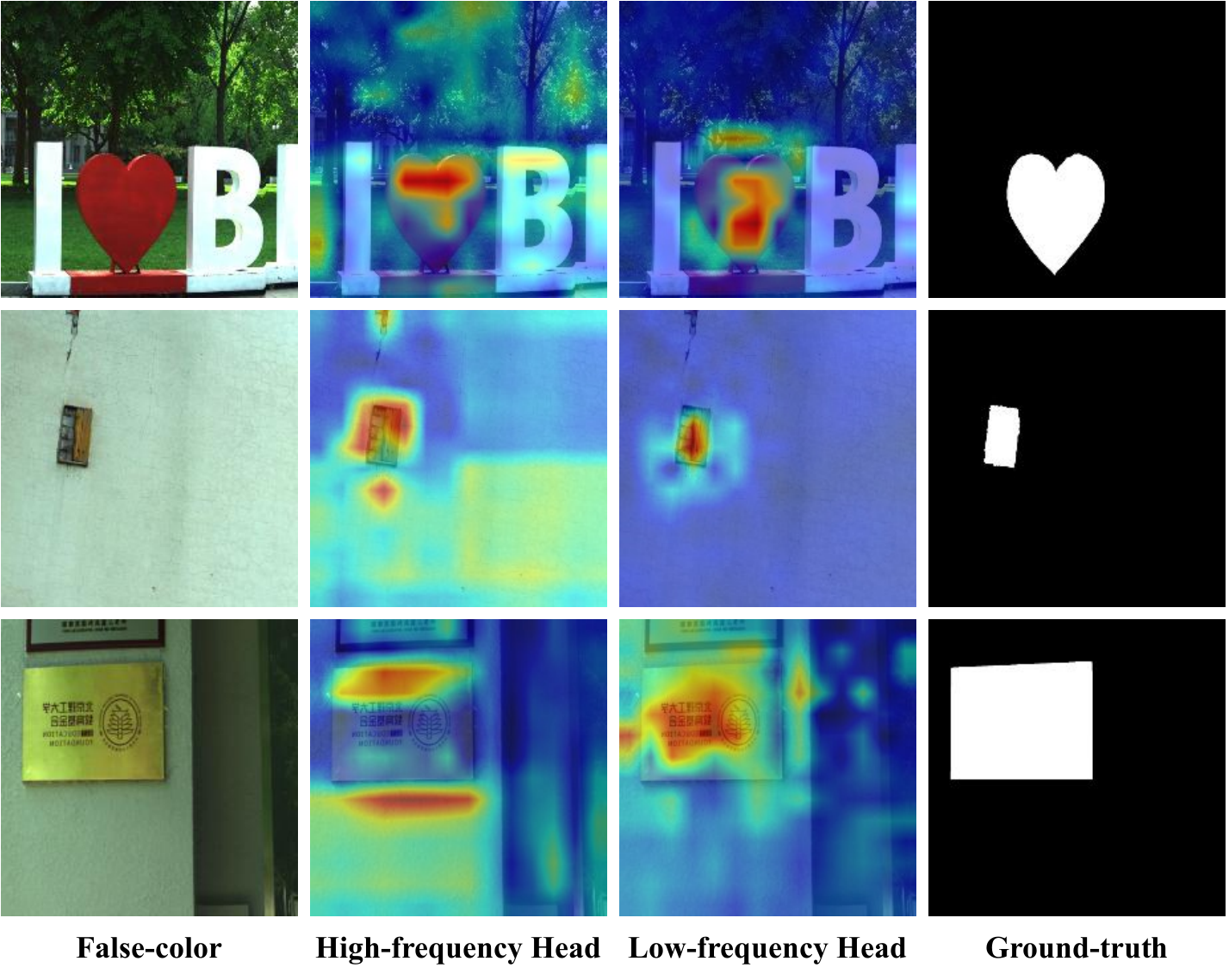} 
        \caption {Visualization of the output of features by the high-frequency attention head and the low-frequency attention head. The former attends to the edge of salient objects, while the latter focuses more on salient objects.}
    \label{middle_feats}
\end{figure}

\nbf{Visualization of Attention Features}
The output features of the high-frequency attention head and the low-frequency attention head in the Mixed-frequency Attention are displayed in \cref{middle_feats}. It is evident that the features generated by the high-frequency attention head are predominantly concentrated along the edges of salient objects. In contrast, the features produced by the low-frequency attention head are primarily focused on the salient objects themselves.

\begin{figure*}[!ht]
    \centering
    \includegraphics[width=\linewidth]{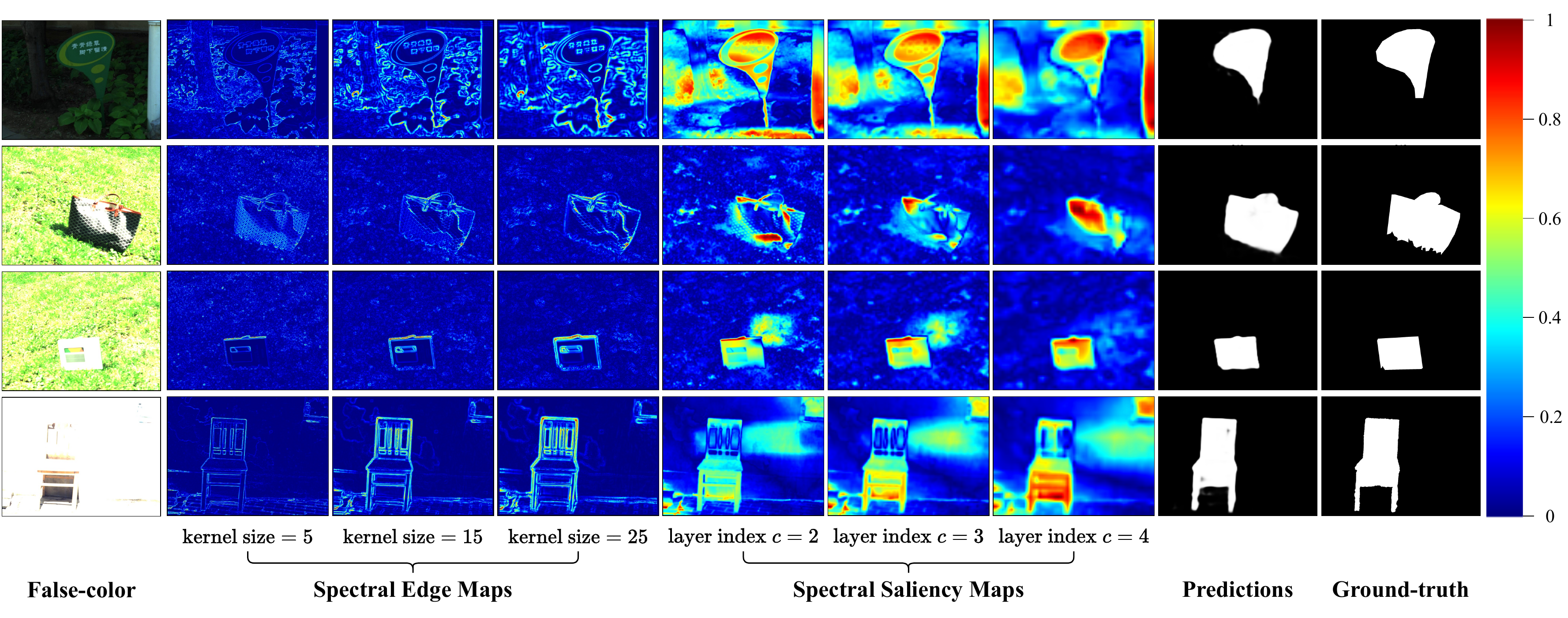}
        \caption {Visualization of Spectral Edge and Spectral Saliency maps from SEO and SSG. The variation of gradient kernel sizes and the layer index $c$ of the ``center" pixel result in different Spectral Edge and Spectral Saliency maps, respectively.}
    \label{fig:edge_and_SS}
\end{figure*}

\nbf{Visualization of SEO and SSG Output}
The visualization of the Spectral Saliency and Spectral Edge maps can be observed in \cref{fig:edge_and_SS}. It is worth noting that the choice of kernel size significantly influences the resulting edge features. Employing a smaller gradient convolution kernel yields a more detailed edge image, as depicted in the leftmost edge map. Conversely, utilizing a larger gradient convolution kernel leads to a clearer overall contour of the object, as depicted in the rightmost edge map.
Similarly, the saliency maps obtained from the upper layer pairs in the pyramid (with a smaller layer index $c$) typically encompass complete salient objects. In contrast, the saliency maps obtained from the lower layer pairs (with a larger layer index $c$) offer greater accuracy in capturing salient regions.

\subsection{Results on HS-SOD}
\nbf{Quantitative Results}
The quantitative comparison results on the HS-SOD dataset are presented in \cref{Dingliang}. 
Our SMN, irrespective of the backbone used, achieves notable performance scores, with SMN-P attaining MAE, \(S_\alpha\), \(F_\beta^\text{max}\), AUC, and CC values of \(0.068\), \(0.788\), \(0.723\), \(0.916\), and \(0.718\), respectively. Notably, our method outperforms traditional methods and SUDF across all evaluation metrics on this dataset.
Compared to RGB-image-based methods, SMN-S slightly lags behind in two metrics, namely MAE and $S_\alpha$. However, it outperforms both BASNet and U2Net in the remaining three evaluation metrics. 
SMN-S outperforms U2Net by increasing $F_\beta^\text{max}$ by $6.8\%$ and AUC by $5.3\%$. Similarly, SMN-P outperforms U2Net by increasing $F_\beta^\text{max}$ by $17.2\%$ and AUC by $7.2\%$. These results underscore the overall efficacy and competitiveness of our proposed SMN on the HS-SOD dataset.

\begin{figure}[tp]
    \centering
    \includegraphics[width=\linewidth]{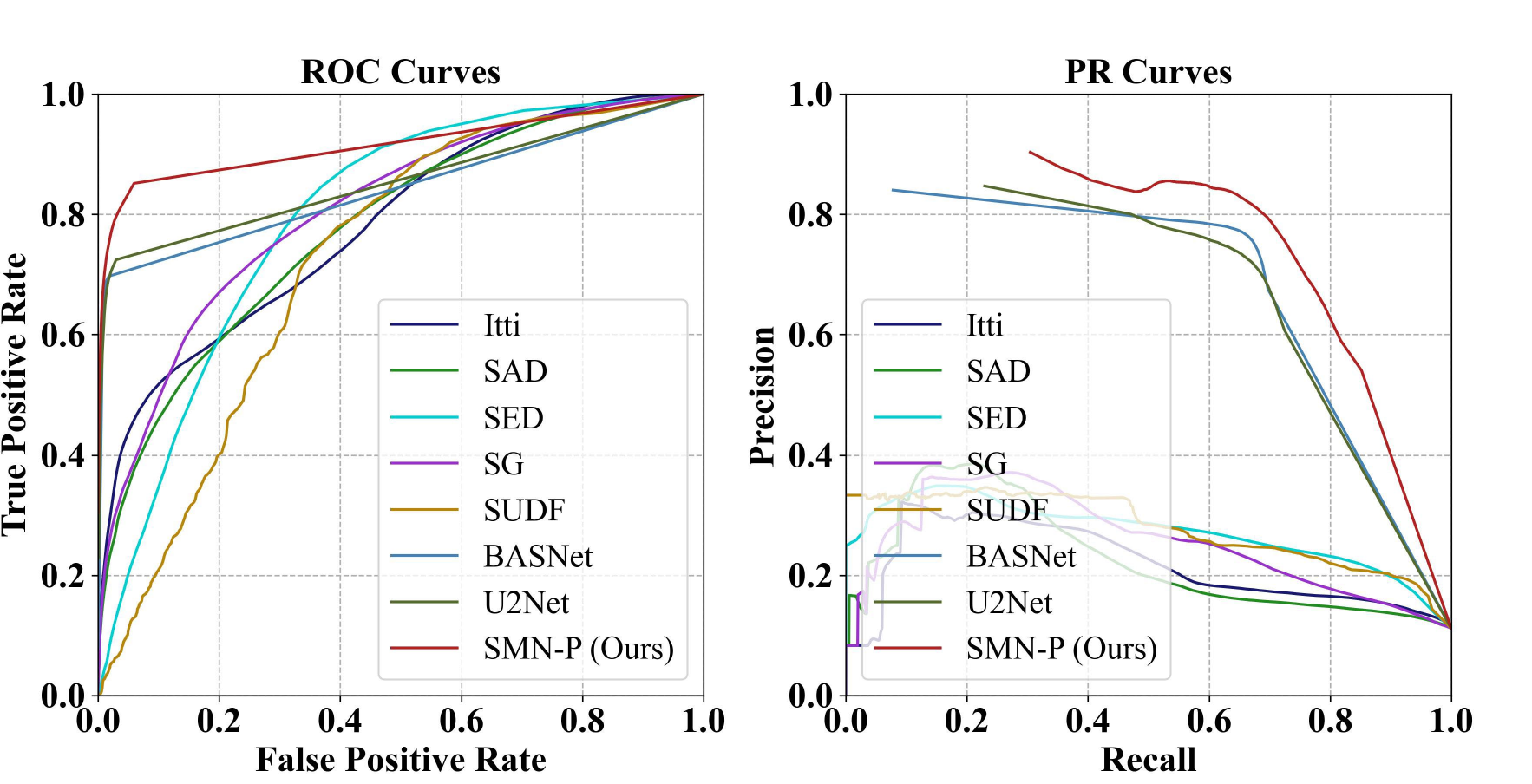}
        \caption {
        Comparison of ROC and PR curves for multiple models on HS-SOD dataset. 
        SMN, represented by a red line, demonstrates a clear advantage in both ROC and PR curves.
        }
    \label{PR curve on HSSOD}
\end{figure}

The comparison of ROC and PR curves between our SMN and other methods on the HS-SOD dataset can be observed in \cref{PR curve on HSSOD}. The ROC curve of SMN, depicted by the red line, demonstrates its proximity to the point (0, 1), indicating a clear advantage over the other methods. Moreover, its advantage in the PR curve is also obvious. By considering the ROC curves, PR curves, and various evaluation metrics, our SMN showcases effectiveness in the context of the HS-SOD dataset.

\begin{figure*}[tp]
    \centering
    \includegraphics[width=\linewidth]{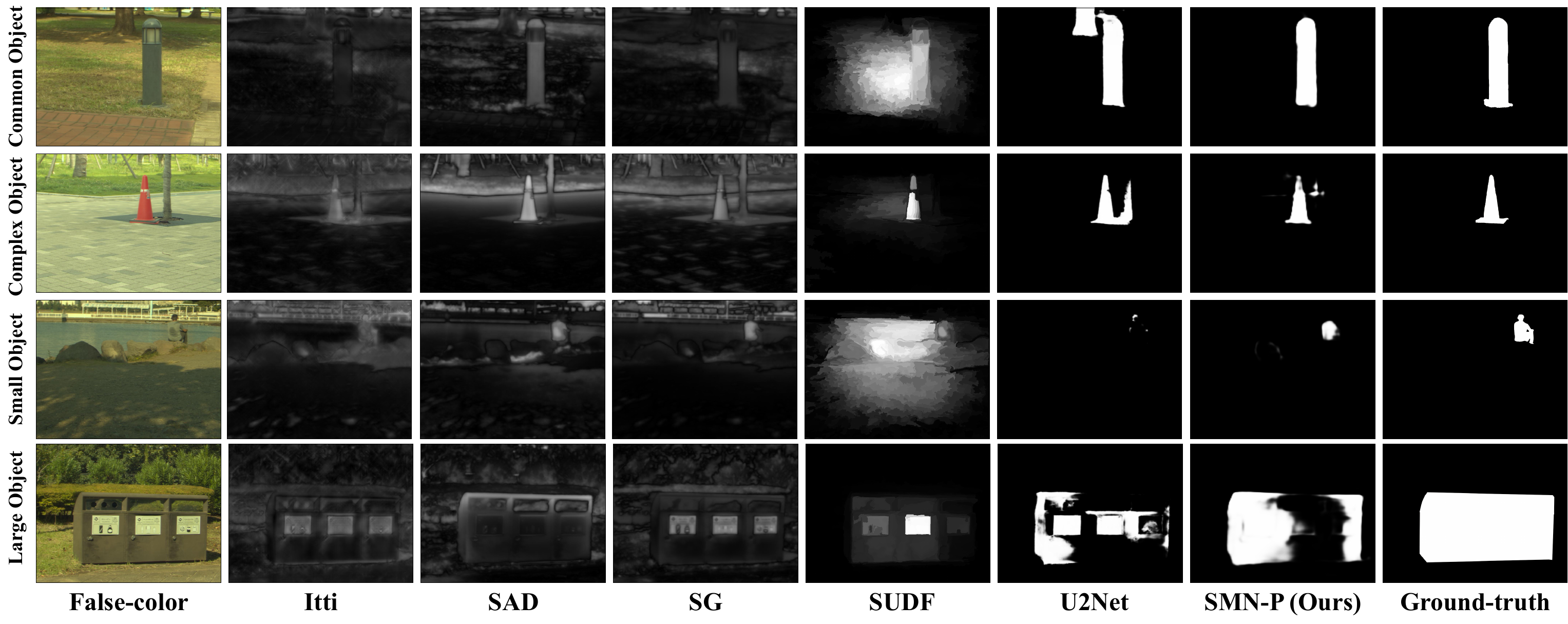}
        \caption {Qualitative Results on HS-SOD dataset. SMN outperforms other methods and is most similar to the ground-truth.}
    \label{fig: HS-SOD result}
\end{figure*}

\nbf{Qualitative Results}
The qualitative results on the HS-SOD dataset are presented in \cref{fig: HS-SOD result}. It can be observed that conventional methods and SUDF exhibit numerous errors and incompleteness in their saliency detection outputs.
In a typical scenario, when compared to U2Net, our SMN demonstrates higher accuracy in detecting objects such as tree trunks and street lamps. This is attributed to the fact that U2Net solely relies on color information for salient object detection and struggles to differentiate objects with similar colors accurately. Conversely, SMN leverages spectral information derived from material properties, enabling it to better distinguish objects with similar colors.
However, in more complex scenes, both SMN and U2Net exhibit errors in their detection results, suggesting the challenges associated with accurate detection.
Regarding small objects, our SMN model exhibits a more comprehensive detection result compared to U2Net. For instance, in the small object scene, SMN successfully identifies the seated person's back as a salient object, whereas U2Net only detects a portion of the person's head.
On the other hand, in the large object scene, both SMN and U2Net face challenges in fully detecting the target objects. In such cases, the performance of both methods is limited.

\subsection{Ablation Study}
\begin{figure*}[htp]
    \centering
    \includegraphics[width=\linewidth]{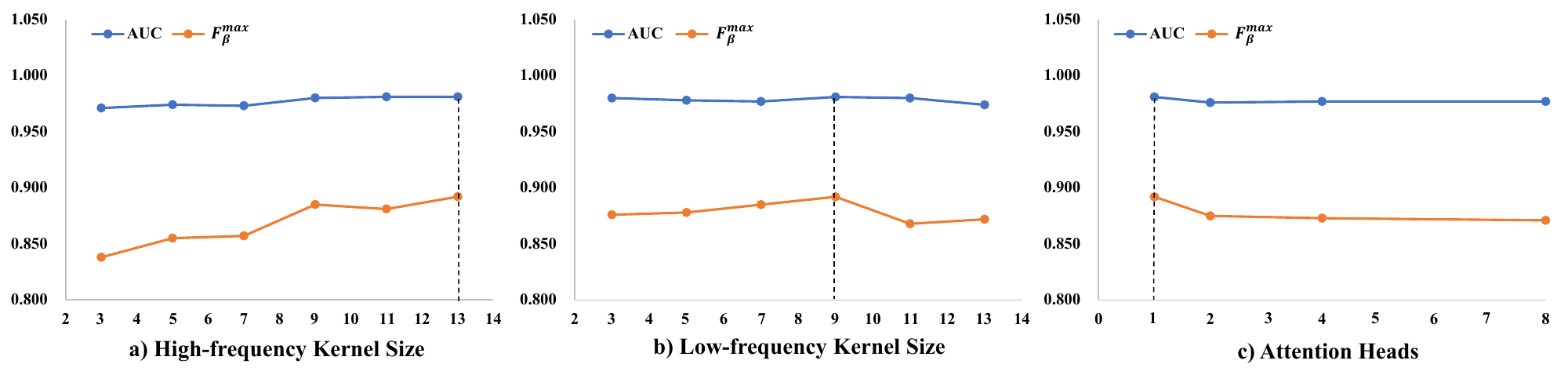} 
        \caption {Hyperparameter analysis of the kernel sizes and number of heads in the neighborhood attention mechanism.}
    \label{hyper analysis}
\end{figure*}

We conducted ablation studies on our HSOD-BIT dataset, choosing PVTv2-b1~\cite{wang2021pvtv2} as the backbone network for the low-frequency embedding.

\nbf{Hyperparameter Analysis}
As previously mentioned, the kernel size of the neighborhood attention mechanism employed in the high-frequency and low-frequency heads differs due to the distinct input and objective of these heads. Hence, we conducted a hyperparameter analysis on the kernel sizes, as well as the number of attention heads. The results, depicted in \cref{hyper analysis}, highlight the significant impact of these hyperparameters on the model's detection performance.
For instance, let us consider the evaluation metrics AUC and $F_\beta^\text{max}$. 
As the kernel size of high-frequency attention increases, the model's detection performance gradually improves, reaching its peak when the kernel size is 13. Further increasing the kernel size may yield better results; however, it is important to note that the neighborhood attention mechanism currently supports a maximum kernel size of 13, limiting any further increase.

Regarding the low-frequency branch, the detection performance of the model achieves its highest value when the kernel size is set to 9, taking into account both evaluation metrics. Furthermore, increasing the number of attention heads has a noticeable negative impact on the model's detection performance.
Based on these findings, we have determined the optimal values for the three hyperparameters: the kernel size for the high-frequency attention is set to 13, the kernel size for the low-frequency attention is set to 9, and each attention mechanism employs a single attention head.

\begin{table}[tp]
    \centering
    \caption{Ablation Study of Input Data.}
    \label{Input data}
    \begin{tabular}{cccc|cc}
    \toprule
        False-color & HSI & Spec. Edge & Spec. Sal. & $F_\beta^\text{max}$ $\uparrow$ & AUC $\uparrow$\\
    \midrule
        \color{green} \ding{51} & \color{red} \ding{55} & \color{red} \ding{55} & \color{red} \ding{55} & 0.887  & 0.978 \\ 
        \color{red} \ding{55} & \color{green} \ding{51} & \color{red} \ding{55} & \color{red} \ding{55} & 0.867  & 0.967 \\ 
        \color{red} \ding{55} & \color{green} \ding{51} & \color{green} \ding{51} & \color{red} \ding{55} & 0.877  & 0.972 \\ 
        \color{red} \ding{55} & \color{green} \ding{51} & \color{red} \ding{55} & \color{green} \ding{51} & 0.881  & 0.979 \\ 
        \color{red} \ding{55} & \color{green} \ding{51} & \color{green} \ding{51} & \color{green} \ding{51} & \textbf{0.892}  & \textbf{0.981}\\ 
    \bottomrule
    \end{tabular}
\end{table}

\nbf{Comparison with Inputting RGB Images}
In order to assess the importance of using HSI as input, we conducted an experiment where we removed two modules specifically designed for HSI, namely SSG and SEO, and directly input false-color images into the SMN. It is worth noting that the hyperparameters of the network remained unchanged throughout this experiment.
As illustrated in \cref{Input data}, the performance of the modified SMN model, measured in terms of $F_\beta^\text{max}$ and AUC, yielded values of $0.887$ and $0.978$, respectively. These values were found to be lower compared to the complete SMN model, which achieved scores of $0.892$ and $0.981$ in the same metrics. This outcome clearly indicates that the simple conversion of HSI to false-color images is less effective in the context of salient object detection, emphasizing the necessity of utilizing HSI as input for achieving superior performance.

\nbf{Usefulness of Two Plug-and-play Operators}
To assess the effectiveness of the plug-and-play modules, SEO and SSG, we conducted experiments where we removed each module individually and compared the results with the baseline.

The baseline experiment involved inputting the complete HSI into the SMN without any modifications. 
When both SEO and SSG modules were removed, the first convolutional layer in the frequency-specified embeddings was randomly initialized, and the input channels were changed to 50. This configuration resulted in the poorest detection performance, with an $F_\beta^\text{max}$ score of only $0.867$ and an AUC of $0.967$.
This outcome highlights the usefulness of the SEO and SSG modules in improving the detection performance of the SMN, underscoring their significance in the context of HSOD.

\noindent \textit{Effect of SEO.}
By incorporating the SEO module, we made modifications to the inputs of the high-frequency and low-frequency embeddings. Specifically, the Spectral Edge map was used as the input for the high-frequency embedding, while the original HSI was retained as the input for the low-frequency embedding.
A comparison between the second row and the third row of \cref{Input data} reveals notable improvements in $F_\beta^\text{max}$ and AUC, indicating a significant enhancement in the detection performance. These results serve as evidence supporting the effectiveness of the SEO module.

\noindent \textit{Efficacy of SSG.}
Upon integrating the SSG module into the baseline configuration, the input for SMN consists of the complete HSI and the Spectral Saliency map. The results depicted in \cref{Input data} exhibit noticeable improvements in both the $F_\beta^\text{max}$ and AUC metrics, thereby affirming the effectiveness of the SSG module. Furthermore, as the Spectral Saliency map provides valuable insights into the approximate location of salient objects, it serves as a crucial information source for SMN's saliency detection. Consequently, the inclusion of the SSG module yields more substantial enhancements in the detection performance compared to solely employing SEO.

\noindent \textit{Impact of SEO and SSG.}
The final row of \cref{Input data} demonstrates that the combined utilization of the SEO and SSG modules, which convert the HSI into edge images and saliency maps, respectively, produces the most remarkable detection performance. Notably, the simultaneous application of both modules yields a more substantial enhancement in detection performance compared to employing either module individually. This outcome can be attributed to the fact that the SEO and SSG modules effectively transform the HSI into edge images and saliency maps, respectively. These transformed representations provide more accurate and suitable high-frequency and low-frequency information for SMN, aligning with the requirements of our specially designed model.

\begin{table}[tp]
    \centering
    \caption{The Effectiveness of Mixed-frequency Attention Module, Shallow Feature, and High-frequency Information.}
    \label{effect of component}
    \begin{tabular}{l|ccccc}
    \toprule
        Models & MAE $\downarrow$ & $F_\beta^\text{max}$ $\uparrow$ & $S_\alpha$ $\uparrow$ & AUC $\uparrow$ & CC $\uparrow$ \\ 
        \midrule
        SMN \textit{w/o} MA & 0.036  & 0.862  & 0.878  & 0.978  & 0.863 \\ 
        SMN \textit{w/o} Sha. Fea. & 0.040  & 0.848  & 0.868  & 0.970  & 0.845 \\ 
        SMN \textit{w/o} HF & \textbf{0.033} & 0.871 & \textbf{0.887} & 0.976 & 0.871 \\
        SMN & 0.034  & \textbf{0.892}  & 0.872  & \textbf{0.981}  & \textbf{0.874}  \\
    \bottomrule
    \end{tabular}
\end{table}

\nbf{Effect of Mixed-frequency Attention Module}
We investigate the impact of the Mixed-frequency Attention (MA) module. An alternative feature fusion approach involved concatenating deep edge information and saliency information along the channel dimension while also modifying the input channel number of the first convolutional layer in the Saliency-edge-aware Decoder. This experimental setup is denoted as SMN \textit{w/o} MA.

By comparing the results in the first and last rows of \cref{effect of component}, it becomes evident that the inclusion of MA has a significant positive effect on the detection performance of the model. MA facilitates the self-refinement of low-frequency information through the utilization of a self-attention mechanism, enabling it to concentrate more on salient objects. Furthermore, the cross-attention mechanism promotes interaction between high and low-frequency features, leading to the generation of more accurate low-frequency features within the constraints imposed by high-frequency information.

\nbf{Effect of Shallow Feature} 
To investigate the influence of shallow features in the Saliency-edge-aware Decoder, we conducted an experiment where these features were eliminated, resulting in a conventional decoder. This experimental setup is referred to as SMN \textit{w/o} Sha. Fea.
Upon examining the results presented in \cref{effect of component}, it becomes apparent that the inclusion of shallow features during the decoding process significantly enhances the detection performance. As the model progresses deeper into the network, the spatial dimensions of the features gradually decrease, leading to a loss of fine details. When decoding is performed solely based on these deep features, the resulting outcomes become less precise. Therefore, integrating shallow features from the encoder at the decoding stage is crucial to compensate for the loss of intricate information and improve the overall detection performance.

\nbf{Impact of High-frequency Information}
To investigate the role of high-frequency information in salient object detection, we conducted an experiment where we removed the high-frequency inputs and solely relied on low-frequency information. This experimental setup involved generating a Spectral Saliency map from the HSI using the SSG module and converting it into low-frequency features through low-frequency embeddings. Subsequently, these features underwent self-attention in the MA for self-refinement and were decoded to obtain the final saliency map without including shallow edge information as input to the decoder. This experiment is denoted as SMN \textit{w/o} HF.
By comparing the last two lines in \cref{effect of component}, it becomes evident that including high-frequency information is crucial for achieving robust salient object detection performance in the SMN model. The results demonstrate the necessity of incorporating inputs from both high and low frequencies for effective detection.

\subsection{Extensive Experiment on RGBT SOD}
Thermal images possess the capability to effectively capture temperature information pertaining to objects within a given scene. In these images, objects exhibiting higher temperatures exhibit greater intensity, thus distinguishing themselves. This particular characteristic bears a resemblance to our Spectral Saliency map. As a result, our proposed SMN is employed in RGBT datasets with the purpose of assessing and confirming the generalizability of our proposed methodology.

\nbf{Experimental Settings}
Given the alterations in the task and dataset, we adjusted the batch size to 32 and set the maximum learning rate to $1 \times 10^{-2}$. The backbone network is PVTv2-b1.

\nbf{Datasets}
Our training and test sets are consistent with the dataset used by Tu \etal~\cite{Cong_2022_Does}.
To provide a comprehensive comparison, we compare the performance of SMN with several existing methods, including MTMR~\cite{Wang_2018_RGBT}, EGNet~\cite{Zhao_2019_EGNet}, CPD~\cite{Wu_2019_Cascaded}, BASNet~\cite{Qin_2019_BASNet}, and ADF~\cite{Tu_2022_RGBT}. Evaluation of the methods is conducted using two metrics: $F_\beta^\text{max}$ and MAE.

\nbf{Quantitaive Results}
As depicted in \cref{tab: RGBT exps}, the SMN demonstrates commendable detection outcomes, exhibiting the lowest MAE values across all three datasets, namely $0.043$, $0.027$, and $0.044$, respectively. Moreover, on the $F_\beta^\text{max}$ metric, SMN outperforms ADF on both the VT821 and VT5000 datasets and only lags behind ADF by $0.012$ on the VT1000 dataset.
During the RGBT SOD experiment, the SSG and SEO modules are excluded. These modules, integral to processing hyperspectral data, extract edge and saliency information from a spectral standpoint. Nonetheless, in the RGBT SOD experiment, solely RGB images are employed for edge extraction, leading to an inherent loss of information. Additionally, while our Spectral Saliency maps encompass a collection of spectral saliency images, there exists only a single thermal image, thereby providing comparatively less information. Despite these, SMN's detection performance remains robust, amply demonstrating the effectiveness and superiority of our proposed SMN.

\begin{figure*}
    \centering
    \includegraphics[width=\linewidth]{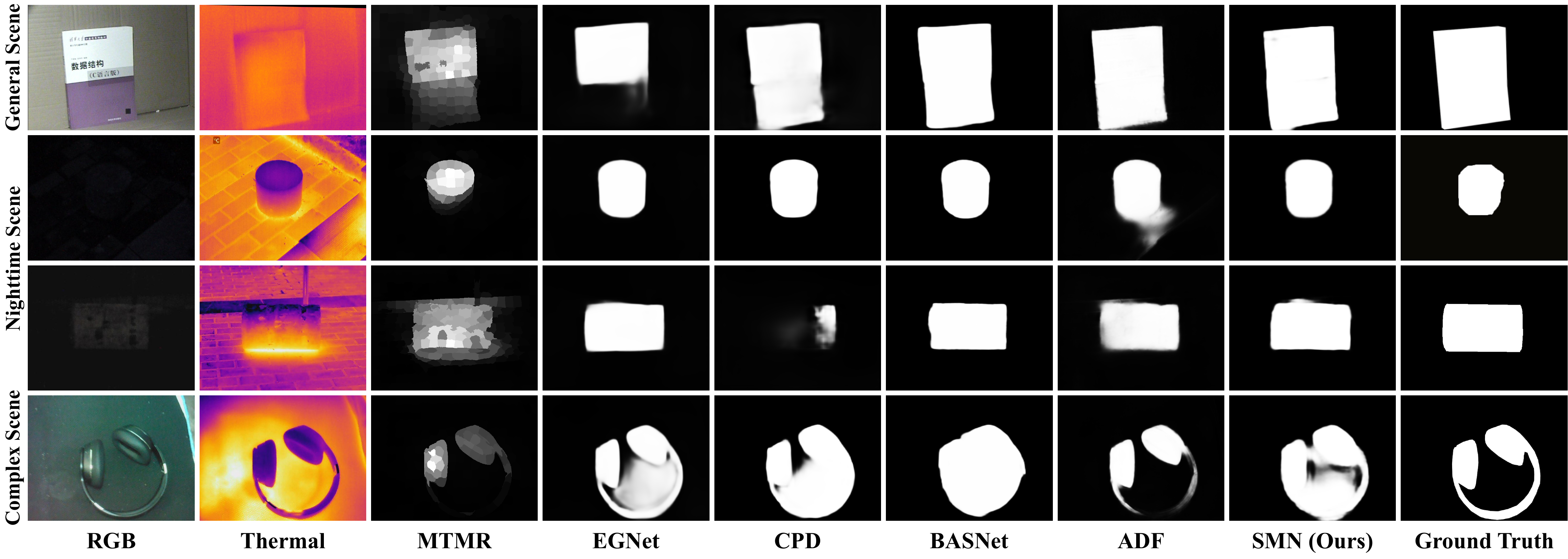}
        \caption{Qualitative results of RGBT SOD. Our SMN achieves satisfactory detection results.}
    \label{fig: RGBT}
\end{figure*}

\begin{table}
\centering
\caption{Results for RGBT SOD on VT821, VT1000 and VT5000 datasets.}
\label{tab: RGBT exps}
\setlength{\tabcolsep}{1.5mm}{
\begin{tabular}{l|cc|cc|cc} 
\toprule
\multirow{2}{*}{Methods} & \multicolumn{2}{c|}{VT821} & \multicolumn{2}{c|}{VT1000} & \multicolumn{2}{c}{VT5000} \\ \cmidrule{2-7} 
                        & $F_\beta^\text{max}$ $\uparrow$        & MAE $\downarrow$         & $F_\beta^\text{max}$ $\uparrow$        & MAE $\downarrow$          & $F_\beta^\text{max}$ $\uparrow$        & MAE $\downarrow$         \\ \midrule
MTMR~\cite{Wang_2018_RGBT}                    & 0.747        & 0.108       & 0.754        & 0.119        & 0.662        & 0.115       \\
EGNet~\cite{Zhao_2019_EGNet}                   & 0.795        & 0.063       & 0.917        & 0.033        & 0.839        & 0.051       \\
CPD~\cite{Wu_2019_Cascaded}                     & 0.786        & 0.079       & 0.914        & 0.031        & 0.847        & 0.047       \\
BASNet~\cite{Qin_2019_BASNet}                  & 0.803        & 0.067       & 0.913        & 0.030        & 0.820        & 0.055       \\
ADF~\cite{Tu_2022_RGBT}                     & 0.804        & 0.077       & \textbf{0.923}        & 0.034        & 0.863        & 0.048       \\ \midrule
SMN (Ours)                     & \textbf{0.831}       & \textbf{0.043}       & 0.911        & \textbf{0.027}        & \textbf{0.908}        & \textbf{0.044}      \\
\bottomrule
\end{tabular}
\vspace{0.1cm}
}
\end{table}

\nbf{Qualitative Results}
The qualitative results of RGBT SOD are visually presented in \cref{fig: RGBT}. Our proposed methods have demonstrated promising outcomes in terms of salient object detection. When compared to MTMR, SMN exhibits enhanced accuracy and completeness in detecting salient objects across diverse scenes. In the general scene, SMN achieves detection results better than MTMR and EGNet. In nighttime scenes, SMN outperforms MTMR, CPD, and ADF by producing more precise detection outcomes with sharper edges. Nonetheless, in the complex scene, SMN exhibits limitations in effectively detecting finer details of head-worn headphones, resulting in relatively weaker performance when compared to ADF.

\section{Conclusion}
\label{sec:conclusion}
In this study, we introduce a novel lightweight model, Spectrum-driven Mixed-frequency Network (SMN), for hyperspectral salient object detection. Our approach is motivated by the insight that spectral information can be leveraged to extract features with two distinct frequencies. To this end, we develop two plug-and-play operators, namely the Spectral Saliency Generator and the Spectral Edge Operator. Furthermore, we design a customized Mixed-frequency Attention module that effectively utilizes the complementarity of these features to generate saliency maps with high-fidelity edges. Experiment results demonstrate our SMN's superiority to state-of-the-art HSOD methods.

Although our method currently surpasses those based on RGB images, the advantage is not substantial. In the future, we plan to construct datasets that more effectively highlight the benefits of utilizing hyperspectral information for saliency detection. Additionally, we aim to further reduce the model size and enhance its speed, facilitating deployment on computation and memory-limited devices.

\section{Acknowledgement}
This work was financially supported by the National Key Scientific Instrument and Equipment Development Project of China (No. 61527802), the National Natural Science Foundation of China (No. 62101032), the Postdoctoral Science Foundation of China (Nos. 2021M690015, 2022T150050), and Beijing Institute of Technology Research Fund Program for Young Scholars (No. 3040011182111).

\bibliographystyle{IEEEtran}
\bibliography{ref}

\end{document}